\documentclass[journal,onecolumn,web]{ieeecolor}
\usepackage{generic}
\usepackage{cite}
\usepackage{amsmath,amssymb,amsfonts}
\usepackage{algorithmic}
\usepackage{graphicx}
\usepackage{subcaption}
\usepackage{algorithm,algorithmic}
\usepackage{hyperref}
\usepackage{cleveref}
\crefname{figure}{fig}{figures}
\Crefname{figure}{Fig}{Figures}
\hypersetup{hidelinks}
\usepackage{textcomp}
\usepackage{comment}
\usepackage{bm}
\usepackage{epstopdf}
\usepackage{epsfig}
\def\BibTeX{{\rm B\kern-.05em{\sc i\kern-.025em b}\kern-.08em
    T\kern-.1667em\lower.7ex\hbox{E}\kern-.125emX}}
\begin{document}
\title{Adaptive Fault-tolerant Control of Underwater Vehicles with Thruster Failures}
\author{Haolin Liu, Shiliang Zhang,~\IEEEmembership{Member, IEEE}, Shangbin Jiao$^*$, Xiaohui Zhang, Xuehui Ma,Yan Yan, Wenchuan Cui, Youmin Zhang,~\IEEEmembership{Fellow, IEEE}
\thanks{$*$Corresponding author.\\This work was supported  by the National Natural Science Foundation of China via the grant No. 62371388, 62127809, and 62103082, and by Scientists \& Engineers Team (No.23KGDW0011 and 2024QCY-KXJ-014), and by the Qin Chuangyuan Innovation Platform (No.23TSPT0002). Haolin Liu, Shangbin Jiao, Xiaohui Zhang,  Xuehui Ma, Yan Yan, and Wenchuan Cui  are with Xi'an University of Technology, Xi'an, China (haolinliu@xaut.edu.cn, jiaoshangbin@xaut.edu.cn, xhzhang@xaut.edu.cn, xuehui.yx@gmail.com, yan.y@xaut.edu.cn, 1220311022@stu.xaut.edu.cn). Shiliang Zhang is with University of Oslo, Norway (shilianz@ifi.uio.no). Youmin Zhang is with Concordia University, Montreal, Canada (youmin.zhang@concordia.ca).}
}
\maketitle
\begin{abstract}
This paper presents a fault-tolerant control for the trajectory tracking of autonomous underwater vehicles (AUVs) against thruster failures. We formulate faults in AUV thrusters as discrete switching events during a UAV mission, and develop a soft-switching approach in facilitating shift of control strategies across fault scenarios. We mathematically define AUV thruster fault scenarios, and develop the fault-tolerant control that captures the fault scenario via Bayesian approach. Particularly, when the AUV fault type switches from one to another, the developed control captures the fault states and maintains the control by a linear quadratic tracking controller. With the captured fault states by Bayesian approach, we derive the control law by aggregating the control outputs for individual fault scenarios weighted by their Bayesian posterior probability. The developed fault-tolerant control works in an adaptive way and guarantees soft-switching across fault scenarios, and requires no complicated fault detection dedicated to different type of faults. The entailed soft-switching ensures stable AUV trajectory tracking when fault type shifts, which otherwise leads to reduced control under hard-switching control strategies. We conduct numerical simulations with diverse AUV thruster fault settings. The results demonstrate that the proposed control can provide smooth transition across thruster failures, and effectively sustain AUV trajectory tracking control in case of thruster failures and failure shifts.
\end{abstract}
\section{Introduction}
\label{sec:introduction}
Autonomous underwater vehicles (AUVs) conduct complex maneuvers and mission reconstructions in underwater environments and have gained increasing attention from the acadmia and industry~\cite{Chaffre_2024,Mokhtari_2024,10901967,DBLP:journals/corr/abs-2410-15837,yang2024hardware,bai2025exploring}. Most AUVs rely on propeller-driven thrusters for mobility, with trajectory tracking being a fundamental capability for autonomous underwater missions~\cite{Chu2024,Chen_2021}. Thruster failures can occur in AUV missions from various fault sources \textit{e.g.}, blade blockage, breakage, and motor and gear damage. Thruster failures are the most common AUV actuator failure that comprise ca. 94.52\% of overall AUV failures~\cite{Liu_2023}. Thruster failures can severely degrade AUV trajectory tracking, leading to reduced tracking accuracy and stability or even failed missions~\cite{Li_2023,Jain2018}.

\begin{figure}
    \centering
    \includegraphics[width=0.5\linewidth]{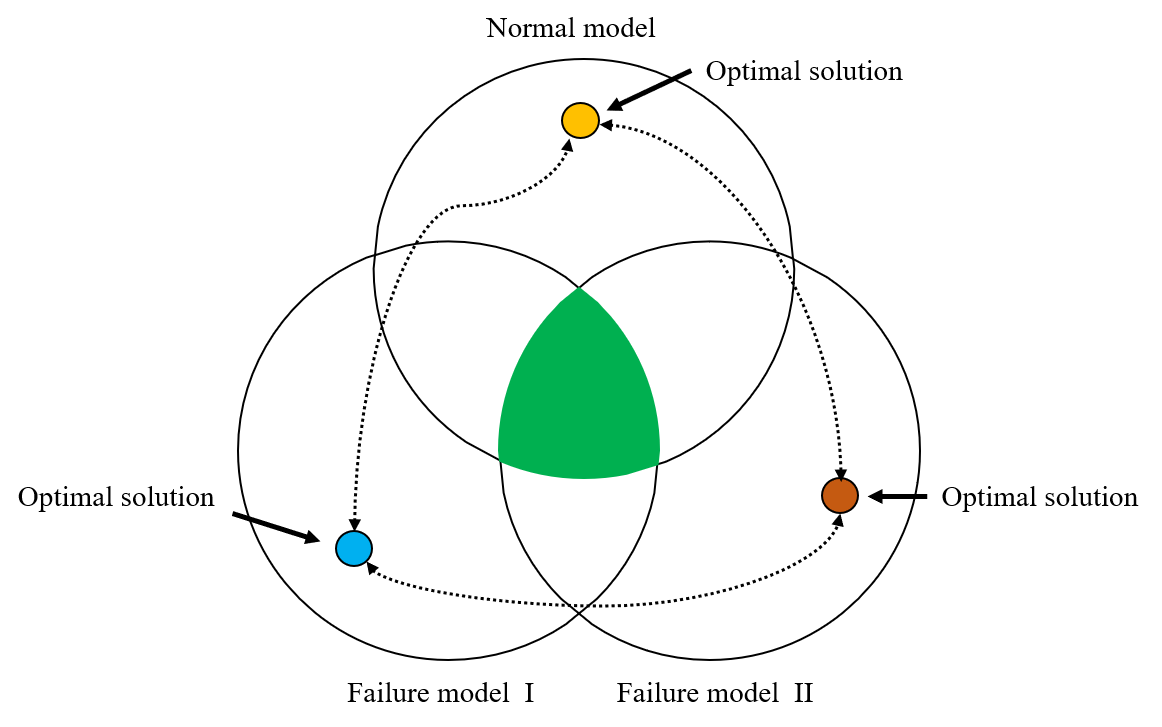}
    \caption{Solution for fault-tolerant control in thrust reconstruction methods. The dotted line represents the transfer of AUV thruster modes. Each of the modes corresponds to a set of feasible solutions for fault-tolerant control. The optimal control solution is determined by the AUV mode and the intersection of modes.}
    \label{fig:enter-label}
\end{figure}

Fault-tolerant control (FTC) for AUVs has been investigated to maintain AUV trajectory tracking despite thruster failures through robust solutions. The design of FTC strategies against thruster failure depends on various factors \textit{e.g.}, AUV geometry, control framework, and the number, location, and redundancy of Thrusters. FTC strategies can be generally categorized into (i) thrust reconstruction, and (ii) adaptive FTC approaches~\cite{Kadiyam_2020}. 
Thrust reconstruction methods, such as sliding mode control \cite{Hao_2019} and linear matrix inequality techniques \cite{Gershon_2019}, consider both normal and fault operational modes in their design. They derive the fault-tolerant control law as the intersection of the feasible solution sets between the nominal and the fault mode, as illustrated in \Cref{fig:enter-label}. 

The problem with thrust reconstruction is that the optimal fault-tolerant control law does not always lie within the intersection of feasible solution sets~\cite{DBLP:journals/jirs/Zogopoulos-Papaliakos21}, and there can the case where there is no feasible solution for the intersection of specific fault modes. Active FTC has emerged as a branch of thrust reconstruction in addressing this issue~\cite{Chen2024,Cerrada2024,Amin_2019}. Active FTC employs Fault Detection and Diagnosis (FDD) and a reconfigurable controller in analyzing and compensating AUV thruster faults. Active FTC adaptively reconfigures and achieves optimal control across different AUV fault modes. Active FTC switches control models according to the detected fault mode~\cite{Gershon_2019}, and the efficiency of the control heavily depends on the fault detection and fault mode switching criteria. As a result, active FTC can frequently experience oscillations and jitters around the switching period. Furthermore, FDD in active FTC can be vulnerable to abnormalities and errors in the detection that adds to the risk of failed control.


Unlike thrust reconstruction approaches, adaptive FTC avoids reliance on precise FDD and estimates the fault status from AUV measurements in real time. 
Gershon \textit{et al.}~\cite{Gershon_2019} employed linear matrix inequalities to isolate the fault and dynamically adapts the controller to a fault-specific fault-tolerant control configuration. Their approach provides enhanced adaptability towards AUV thruster faults, yet this method relies on accurate fault isolation and their control is prone to instability when the fault mode switches. 
Santos \textit{et al.}~\cite{dos_Santos_2016} assumed up to two failed thrusters out of the total four AUV thrusters. They use a deterministic automaton model to determine the failure, and build a dictionary-based control that assigns PID controllers to different thruster failure scenarios. However, their approach is limited in handling AUV faults with complex fault settings and is with restricted generalizability. 
Shi \textit{et al.}~\cite{Shi_2022} proposed a control that can filter out erroneous AUV measurements to improve AUV's tolerance against potential faults and outliers~\cite{DBLP:journals/tnn/ZhangCYZH18,DBLP:journals/cssc/ZhangCYZH19,MA2022157,9189668}. Similarly, Du \textit{et al.}~\cite{Du_2023} developed an adaptive fault-tolerant control against abnormal AUV observations. They predefine fault scenarios and derive control models for normal and abnormal observations, and construct hybrid models for the scenarios that are beyond the predefined ones. Nevertheless, the switching between fault scenarios and the control during the switching are not well analyzed in their design. 
Liang \textit{et al.}~\cite{Liang_2021} adopted support vector machine in classifying underwater vehicle faults, and developed different control strategies \textit{e.g.,} PID, FPID, and MPC~\cite{https://doi.org/10.1155/2017/9402684,7554029,7553664}, for different fault conditions. We emphasize that the transition between thruster fault conditions takes time, and a smooth transition depends on well designed AUV control that supports and stabilizes the switch from one fault type to another. Though fault-tolerant control for AUV thruster failures has been extensively studied, the control for a smooth switch across different faults is not full investigated to underpin the transition period. 


In this paper, we focus on the fault-tolerant control of AUV trajectory tracking, where the AUV is subject to thruster failures. Particularly, we look into the control when the AUV fault transfers from one type to another. To this end, we adopt a Bayesian approach~\cite{yu2024adaptive,MA2022110,ma2022active} that facilitates soft switch between AUV faults for the fault-tolerant control. We use Bayesian method to estimate the AUV thruster fault states, and we derive linear quadratic tracking (LQT) control~\cite{DBLP:journals/tac/MaZLQSH24,DBLP:conf/eucc/MaCZLQS24} according to the estimated fault states. We derive the final control law through aggregating individual LQT controller outputs weighted by their Bayesian posterior probability. Furthermore, we develop the unlocking mechanism against the probability locking of the Bayesian approach in fault state estimation, thus facilitating a fully fledged adaptive fault-tolerant control. We provide rigorous mathematical analysis for our soft switching across failure types and fault-tolerant control against thruster failures, and we demonstrate our design via numerical simulations with detail result reports.


The rest of this paper is organized as follows: Section~\ref{sec-preliminaries} presents the mathematical model of a underwater vehicle equipped with eight thrusters. Section~\ref{sec-controllaw} derives the control law and unlocking strategy using discrete LQT and Bayesian posterior probability weighting. Section~\ref{sec-simulation} evaluates the algorithm's performance through simulations, including scenarios involving two thruster failures and comparisons between soft and hard switching in a horizontal configuration. Finally, Section~\ref{sec-conclusions} concludes the paper.

\section{Preliminaries and Problem Formulation}
\label{sec-preliminaries}
This section lays the fundamentals for modeling the motion of autonomous underwater vehicle (AUV), based on which we formulate the fault-tolerant tracking control problem for AUV. In this section, we take the vehicle model of ``Blue ROV2 Heavy "~\cite{Robotics} when the specifications are necessary in the mathematical analysis for AUV motion. 

\begin{figure}
    \centering
\includegraphics[width=0.7\columnwidth]{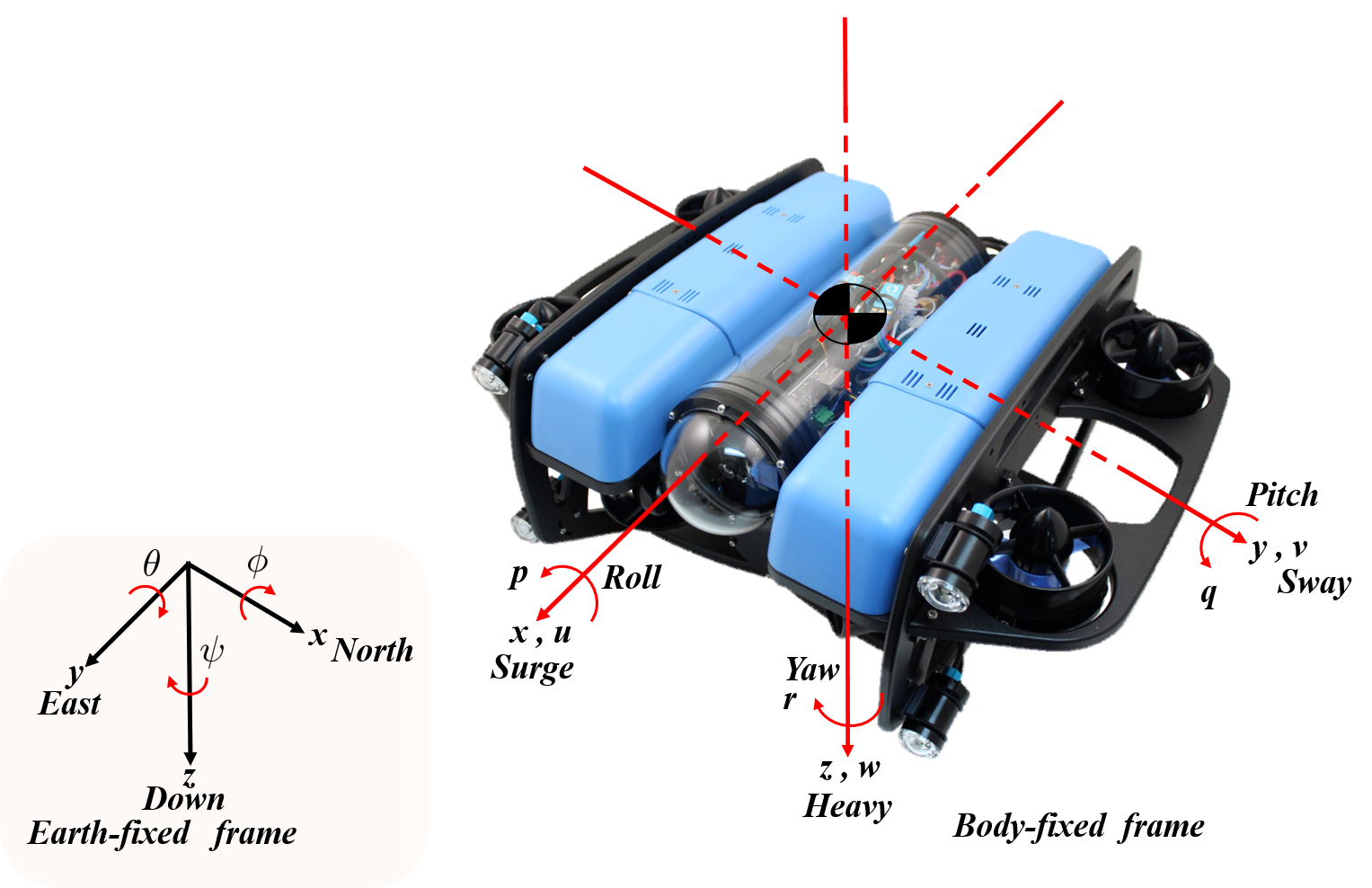 }
    \caption{The two respective coordinate systems of the Blue ROV2 Heavy}
    \label{fig:2}
\end{figure}

\subsection{Dynamics Modeling for AUV}\label{section:auv_dynamics}
This section presents the kinematic model, kinetic model and thruster model for AUV using Fossen's modeling theory~\cite{Fossen_2021} in conjunction with the Newton–Euler method and Lagrange equations\cite{Li_2024}. We use those models to describe AUV motion propelled by thrusters. We define two reference frames: the Earth-fixed frame and the Body-fixed frame as shown in \Cref{fig:2}, for analyzing the AUV motion in six degrees of freedom~\cite{Er2024}. We present the kinematic model for AUV's spatial motion as 
\begin{equation}\label{eq:1}
\dot{\boldsymbol{\eta}}=\bm{J}(\boldsymbol{\eta})\boldsymbol{\nu},
\end{equation}
where $\boldsymbol{\eta}=[x,y,z,\phi,\theta,\psi]^{\mathrm{T}} $ is the position and attitude vector of the AUV with coordinates in the Earth-fixed frame shown in \Cref{fig:2}, $\boldsymbol{\nu}=[u,v,w,p,q,r]^{\mathrm{T}}$ is the linear and angular velocity vector with coordinates in the Body-fixed frame, $\boldsymbol{J}(\boldsymbol{\eta})\in\mathbb{R}^{6\times6}$ is a transformation matrix from the Body-fixed frame to the Earth-fixed frame that 
\begin{equation}
\boldsymbol{J}(\boldsymbol{\eta})=diag[\boldsymbol{J}_{1}(\boldsymbol{\eta}),\boldsymbol{J}_{2}(\boldsymbol{\eta})],
\end{equation}
where the $\boldsymbol{J}_{1}(\boldsymbol{\eta})$ 
is a $(3 \times 3)$ matrix that relates the linear velocity components in the Body-fixed frame $([u, v, w])$ to the Earth-fixed frame $([x, y, z])$. $\boldsymbol{J}_{1}(\boldsymbol{\eta})$ is responsible for the translation transformation and depends primarily on the orientation of the AUV. $\boldsymbol{J}_{2}(\boldsymbol{\eta})$ is a $(3 \times 3)$ matrix that relates the angular velocity components in the Body-fixed frame $([p, q, r])$ to the orientation angles $([\phi, \theta, \psi])$ in the Earth-fixed frame. The definition of $\boldsymbol{J}_{1}(\boldsymbol{\eta})$ and $\boldsymbol{J}_{2}(\boldsymbol{\eta})$ can be found in the literature~\cite{Fossen1995,Fossen2006,Fossen_2021,Antonelli_2021}.

The kinetic model for AUV motion in the Body-fixed frame is \begin{equation}
\label{eq:2}
    \begin{aligned}
\mathbf{M}\dot{\boldsymbol{\nu}}+\mathbf{C}(\boldsymbol{\nu})\boldsymbol{\nu}+\mathbf{D}(\boldsymbol{\nu})\boldsymbol{\nu}+g(\boldsymbol{\eta})=\boldsymbol{\tau},
\end{aligned}
\end{equation}
where $\boldsymbol{M}\in\mathbb{R}^{6\times6}$ is the inertia matrix accounts for the mass and inertia of the AUV taking into consideration the effects of added mass due to the surrounding fluid. $\boldsymbol{C}(\boldsymbol{\nu})\in\mathbb{R}^{6\times6}$ represents the Coriolis-centripetal matrix that arise from the AUV's motion within the fluid taking into account the effects due to added mass. $\boldsymbol{D}(\boldsymbol{\nu})\in\mathbb{R}^{6\times6}$ is the hydrodynamic damping matrix that models the drag forces acting on the AUV due to the fluid, which depends on the velocity and represents resistive forces that oppose the AUV’s motion. $\boldsymbol{g}(\boldsymbol{\eta})\in\mathbb{R}^{6\times1}$ denotes the vector of hydrostatic restoring force and moments, which captures the buoyancy forces and gravitational effects acting on the AUV. $\boldsymbol{\tau}\in\mathbb{R}^{6\times1}$ denotes the control force and moment vector that indicates the inputs from the AUV’s actuators which can be controlled to maneuver the vehicle~\cite{Fossen_2021,Antonelli_2021,Pettinelli2022}.

The thruster model that maps the thruster control input to the ROV's control force and moment is 
\begin{equation}
   \boldsymbol{\tau}=\bm{T}\bm{\Gamma}\bm{u},
    \label{seq:1}
\end{equation}
where $\bm{u}\in \mathbb{R}^{8\times1}$ is the control input vector that dictates the thrust level of the eight thrusters in a ``BlueROV2 Heavey'' AUV. $\bm{T}\in \mathbb{R}^{6\times8}$ is the thrust allocation matrix that maps the forces generated by each thruster to the six degrees of freedom in which the AUV can move. $\Gamma \in \mathbb{R}^{8\times8}$ is the diagonal force coefficient matrix that indicates the effectiveness of each thruster. It represents the relationship between the input signal in $\bm{u}$ and the actual force produced by each thruster. Failures in the thrusters can be reflected in the parameters of the matrix $\Gamma$~\cite{Fossen_2021,Antonelli_2021,Pettinelli2022,Kokegei2016}.  

\subsection{Linear State-space Model for AUV Motion }
This section linearizes the AUV motion models presented in \Cref{section:auv_dynamics}, and transfers the linear model into a linear state-space model. This linearization and simplification lay the basis for the control scheme derivation.

We derive the linear model for AUV motion by linearizing the models in (\ref{eq:1}) and (\ref{eq:2}) with a reference trajectory $\boldsymbol{\nu}_t$, $\boldsymbol{\eta}_t$ and $\boldsymbol{\tau}_t$, and we define the perturbations below to indicate the divergence of the AUV from the reference trajectory:
\begin{equation}
    \begin{gathered}
    \label{eq:3}
\Delta \boldsymbol{\nu} =\boldsymbol{\nu}-\boldsymbol{\nu}_t, \quad
\Delta\boldsymbol{\eta} =\boldsymbol{\eta}-\boldsymbol{\eta}_t, \quad
\Delta\boldsymbol{\tau} =\boldsymbol{\tau}-\boldsymbol{\tau}_t.
\end{gathered}
\end{equation}
Then we introduce the following vector notation
\begin{equation}
    f_{c}(\boldsymbol{\nu})=\boldsymbol{C}(\boldsymbol{\nu})\boldsymbol{\nu},\quad f_{d}(\boldsymbol{\nu})=\boldsymbol{D}(\boldsymbol{\nu})\boldsymbol{\nu}
    \nonumber,
\end{equation}
and we linearize the equation in \eqref{eq:2} and \eqref{eq:1} by
\begin{equation}
   \begin{aligned}
\boldsymbol{M}\Delta\dot{\boldsymbol{\nu}}+\left.\frac{\partial f_c(\boldsymbol{\nu})}{\partial\boldsymbol{\nu}}\right|_{\boldsymbol{\nu}_t}\Delta\boldsymbol{\nu}+\left.\frac{\partial f_d(\boldsymbol{\nu})}{\partial\boldsymbol{\nu}}\right|_{\boldsymbol{\nu}_t}\Delta\boldsymbol{\nu}
   \\+\left.\frac{\partial g(\boldsymbol{\eta})}{\partial\boldsymbol{\eta}}\right|_{\eta_t}\Delta\boldsymbol{\eta}=\Delta\boldsymbol{\tau}
   \end{aligned} 
\end{equation}
and
\begin{equation}
   \begin{aligned}
   \Delta\dot{\boldsymbol{\eta}} = \left.\frac{\partial J_{\boldsymbol{\nu}\boldsymbol{\eta}}(\boldsymbol{\nu},\boldsymbol{\eta})}{\partial \nu}\right|_{\boldsymbol{\nu}_t}\Delta\boldsymbol{\nu} + \left.\frac{\partial J_{\boldsymbol{\nu}\boldsymbol{\eta}}(\boldsymbol{\nu},\boldsymbol{\eta})}{\partial \boldsymbol{\eta}}\right|_{\boldsymbol{\eta}_t}\Delta\boldsymbol{\eta}
   \end{aligned} 
\end{equation}
Defining $\bm{x}_1=\Delta\boldsymbol{\nu}$ and $\bm{x}_2=\Delta\boldsymbol{\eta}$ yields the following linear time-varying model
 \begin{equation}
     \begin{aligned}\label{linearM1}
     &\bm{M} \dot{\bm{x}}_{1}+\boldsymbol{C}(t) \bm{x}_{1}+\boldsymbol{D}(t) \bm{x}_{1}+\boldsymbol{G}(t) \bm{x}_{2} = \boldsymbol{\tau}, \\
     &\dot{\bm{x}}_{2}=\boldsymbol{J}_1(t) \bm{x}_{1}+\boldsymbol{J}_2(t) \bm{x}_{2}, 
     \end{aligned}
 \end{equation}\\
 where\\
 \begin{equation}
  \begin{aligned}
     &\boldsymbol{C}(t)=\frac{\partial f_{c}(\bm{\nu})}{\partial\bm{\nu}}\Bigg|_{\bm{\nu}_t}, \quad \bm{D}(t)=\frac{\partial f_{d}(\bm{\nu})}{\partial\bm{\nu}}\Bigg|_{\bm{\nu}_t}, \\
     & \boldsymbol{G}(t)=\frac{\partial g(\bm{\eta})}{\partial\bm{\eta}}\Bigg|_{\bm{\eta}_t}, \quad \boldsymbol{J}_1(t)=\left.\frac{\partial J_{\bm{\nu}\bm{\eta}}(\bm{\nu},\bm{\eta})}{\partial \bm{\nu}}\right|_{\bm{\nu}_t}=\boldsymbol{J}(\bm{\eta}_t), \\
     &\boldsymbol{J}_2(t)= \left.\frac{\partial J_{\bm{\nu}\bm{\eta}}(\bm{\nu},\bm{\eta})}{\partial \bm{\eta}}\right|_{\bm{\eta}_t}=\frac{\partial \boldsymbol{J}(\bm{\eta})}{\partial\bm{\eta}}\Bigg|_{\bm{\eta}_t}\bm{\nu}_t.
  \end{aligned}
  \nonumber
 \end{equation}
Substitute the thruster model in \eqref{seq:1} into \eqref{linearM1}, and we can derive the following linear state-space model for AUV
\begin{equation} 
\begin{aligned}
\begin{bmatrix}\dot{\bm{x}}_1\\\dot{\bm{x}}_2\end{bmatrix}=&\begin{bmatrix}-\boldsymbol{M}^{-1}[\boldsymbol{C}(t)+\boldsymbol{D}(t)]&-\boldsymbol{M}^{-1}\boldsymbol{G}(t)\\\boldsymbol{J}_1(t)&\boldsymbol{J}_2(t)\end{bmatrix}\begin{bmatrix}\bm{x}_1\\\bm{x}_2\end{bmatrix}\\
&+\begin{bmatrix}\boldsymbol{M}^{-1}\\0\end{bmatrix}\bm{T}\Gamma \bm{u}\\
 \label{eq:8}
\end{aligned}
\end{equation}

Equation~(\ref{eq:8}) presents a continuous-time system model, which is preferred in theoretical analysis. However, it is necessary to discretize the continuous system in practice especially in embedded system applications to facilitate the implementation of real-time signal processing in the hardware. Below we describe the discretization of the continuous-time system.

We first approximate the continuous-time system using the forward Euler method. Assume that the continuous system is sampled with a short sampling interval $\Delta T$, we can discretize the system dynamics as follows:
\begin{equation}
 {\bm{x}}(k+1)=\bm{A}(k)\bm{x}(k)+\bm{B}(k,\Gamma)\bm{u}(k),
 \label{teq:1}
\end{equation}
where the definition of system state $\bm{x}(k)$, and state matrices $\bm{A}$ and $\bm{B}(k,\Gamma)$ are
\begin{equation}
    \bm{x}(k)=[\bm{x}^T_1(k),\bm{x}^T_2(k)]^T,
    \nonumber
\end{equation}
\begin{equation}
    \bm{A}(k) =\bm{I}+ \Delta T\begin{bmatrix}-\boldsymbol{M}^{-1}[\boldsymbol{C}(k)+\boldsymbol{D}(k)]&-\boldsymbol{M}^{-1}\boldsymbol{G}(k)\\\boldsymbol{J}_1(k)&\boldsymbol{J}_2(k)\end{bmatrix},
     \nonumber
\end{equation}
\begin{equation}
    \bm{B}(k,\Gamma) = \Delta T\begin{bmatrix}\boldsymbol{M}^{-1}\\0\end{bmatrix}\bm{T}\Gamma,
     \nonumber
\end{equation}
\begin{equation}
   k=0,1,2,3,\ldots
      \nonumber
\end{equation}
The matrix $\mathbf{A}$ represents the state matrix of the AUV, which is influenced by the geometric structure parameters and the thruster layout of the AUV. Matrix $\mathbf{B}$ represents the input matrix, which depends on the position, orientation, and thrust of the AUV. With the discretized system in \eqref{teq:1}, we gain the foundation for the design and tuning of control laws that can be tailored to specific requirements and constraints in underwater environments. In the following section, we will formulate the fault-tolerant control problem for AUV based on the derived discretized AUV motion model. 

\subsection{Fault-tolerant Control Problem Formulation}
The model presented in \eqref{teq:1} does not take into account uncertainties in the system and potential faults induced by the uncertainties. However, uncertainties inevitably exist in AUV applications that affects the control of AUV and leads to fault in the control to different levels. \textit{E.g.}, sensors are empolyed to measure AUV velocity and position for the estimation of AUV system state, while the sensors introduce measurement noises. There are also inherent uncertainties due to the usage and manufacturing of AUV hardware, \textit{e.g.}, accidents in wiring connections between AUV components, imprecise layout of electronic circuits, and also uncontrollable factors in the environment like temperature fluctuations and electromagnetic interferences. Such uncertainties can render failures in AUV components and reduced or even failed AUV control.

The risk of faults necessitates the integration of fault representation in the model for AUV control. In this paper, we focus on the failures in AUV thrusters, and we revise the AUV model in \eqref{teq:1} to include the impact of faults in AUV thrusters, as shown below:
\begin{equation}
\begin{aligned}
    \bm{x}(k+1)&=\bm{A}(k)\bm{x}(k)+\bm{B}(k,\Gamma)\bm{u}(k)+\omega(k),\\
    \bm{y}(k)&=\bm{H}\bm{x}(k)+v(k),\\
    \Gamma \in \Omega &\triangleq\big \{\Gamma_1, \Gamma_2,\ldots,\Gamma_n\big \},
\end{aligned}
   \label{Teq:2} 
\end{equation}
with the cost  function for the control of AUV as
\begin{equation}\label{Teq:cost_func} 
    J=\frac{1}{2}\mathbb{E}\left[\bm{x}^T_NP\bm{x}_N+\sum^{N-1}_{k=0}(\bm{x}_k^TQ\bm{x}_k+\bm{u}_kR\bm{u}_k)\right],
\end{equation}
where $x(k)$ is the motion state and $y(k)$ represents the measurement at time $k$, $\omega(k)$ is the process noise, $\nu(k)$ is the measurement noise. Both $\omega(k)$ and $\nu(k)$ are zero mean white Gaussian noise with covariances:
\begin{equation}
\begin{aligned}
   \bm{E}{\{\omega(k)\omega^T(k)\}}\sim\boldsymbol{Q}(k)\\
\bm{E}{\{v(k)v^T(k)\}}\sim\boldsymbol{R}(k).
\end{aligned}
\end{equation}
The observation matrix $\bm{H}$ is a full-rank matrix with certain sparsity. $\Gamma$ represents the failure parameter, and $\Omega$ is the bank of models which contains all the mathematical models for AUV thruster failures. We assign the weights of $P \geq 0,Q \geq 0,$and $R \geq 0$ in the cost function \eqref{Teq:cost_func} as the terminal cost weight, state weight and input control weight, respectively. 

Note that in this model, the input matrix $\mathbf{B}$ is related to the thrusters, and the thruster failure parameter $\Gamma$ is therefore integrated in the matrix $\mathbf{B}$. The identification of the value $\Gamma$ from $\Omega$ determines the failure type and thus the resulted AUV motion model and the AUV control strategy. 

There is a problem with the new model when the value of $\Gamma$ switches, indicating the switch of thruster failure types in AUV operation. \textit{I.e.}, the switch in $\Gamma$ leads to abrupt shift in AUV model and AUV control strategy, which can induce vibrations in AUV motion and degraded AUV control. This issue necessitate a ``soft switching'' that ensures stable control across failure types. 

Another issue with the switch of $\Gamma$ is that, the identification of the value of $\Gamma$ can be locked to an existing failure type, neglecting new failure(s) adding to the existing one. This happens since the recognition of failures generally relies on probability approaches, where the identified failure type with the highest probability determines the value of $\Gamma$. This issue can lead to incorrect failure parameters in the AUV model thus failure in AUV control.

In the next section, we detail our fault tolerant control solution where we address the analyzed issues with the switch in the failure parameter $\Gamma$.

\section{Fault Tolerant Control Design}\label{sec-controllaw}
This section details the design of fault tolerant control against AUV thruster failures. We start with an overview of the designed control in \Cref{section:control_overview}, followed by the description of the control law derivation, system state estimation, failure coefficient identification, and failure identification unlocking. Specifically, we derive the optimal control law using Linear Quadratic Tracking (LQT) method for a given failure parameter $\Gamma_s\in \Omega$ in the model \eqref{Teq:2} in \Cref{section:LQT}. Then we provide soft switch for the failure parameter $\Gamma_s$ by the estimation of AUV state and the derivation of posterior probability for the value of $\Gamma_s$ in \Cref{EKF} and \Cref{section:posterior_probability_update}, respectively. Finally, we provide the solution for the locking of failure parameter identification in \Cref{section:unlock_strategy}.

\subsection{Overview of the Control Design}\label{section:control_overview}
 
\begin{figure}
    \centering
    \includegraphics[width=0.7\columnwidth]{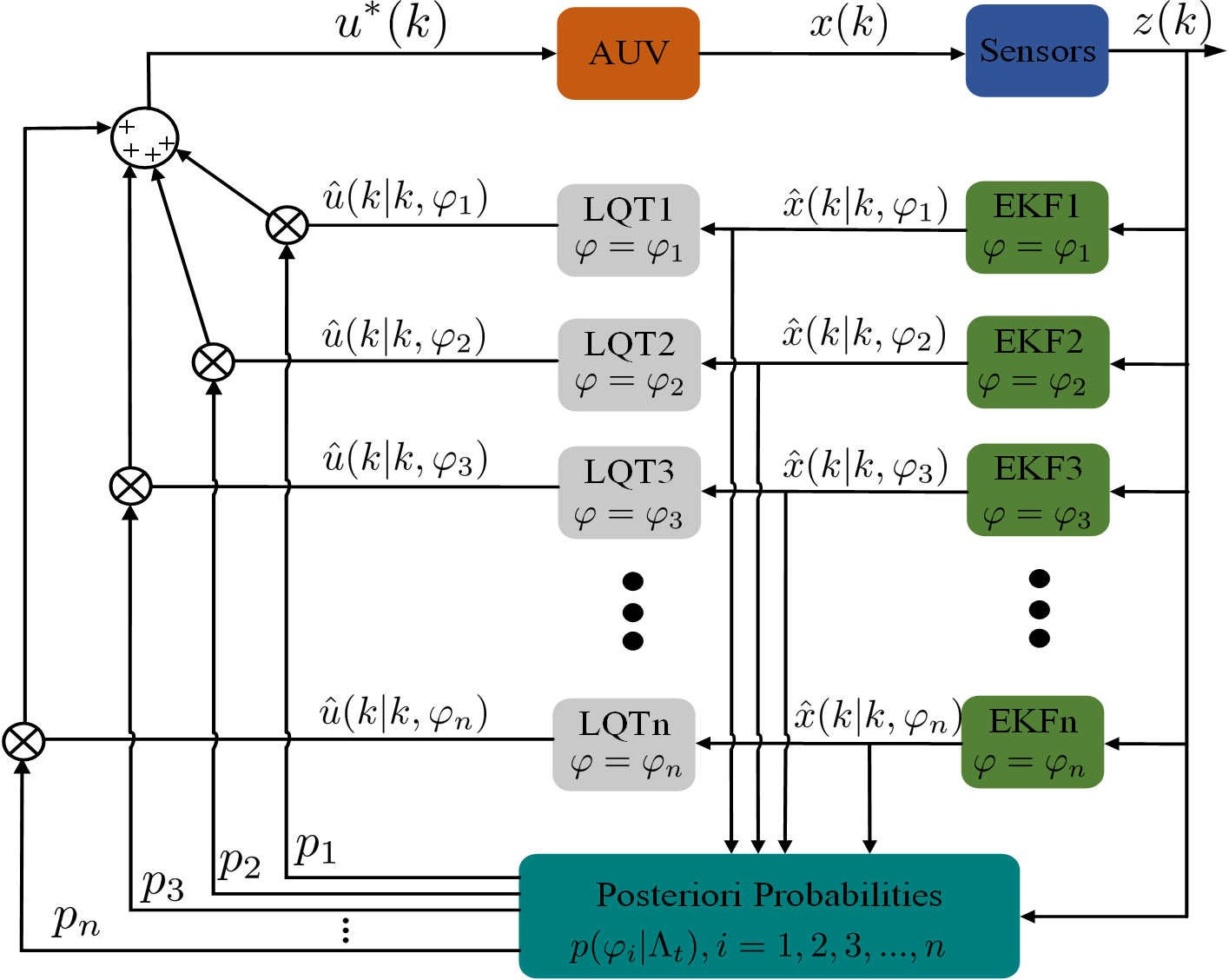  }
    \caption{Structure of the proposed adaptive fault tolerant control (LQT: Linear quadratic tracking. EKF: Extended Kalman Filter. $u$ represents the control signal, $k$ is the time instant in the control, $p_{i}$ is the posteriori probability at time instant $i$, $\Lambda_k$ is the historical observations from the AUV.)}
    \label{fig:1}
\end{figure}

\Cref{fig:1} presents the adaptive fault-tolerant control design, which integrates model-based control and Bayesian probability in addressing thruster failures in AUV control. We derive the model-based control upon AUV dynamics in Section~\ref{sec-preliminaries}, and we use Bayesian probability to scaffold the identification of AUV thruster failures that can lead to soft switch for AUV state estimation. Particularly, we derive the \emph{posterior} probability for different AUV failures, and then use the probability as the weight for the control laws corresponding to individual AUV failure models.

For AUV trajectory tracking subject to thruster failures, we propose an adaptive fault-tolerant control law. We develop the control where the AUV is anticipated to track a given trajectory. We adopt extended Kalman filter in estimating AUV failure state, and we design a model space $\Omega_{\varphi}$ to represent various thruster failure scenarios, with a corresponding control law designed for each $\varphi$. The value of $\varphi$ is identified iteratively by the \emph{posteriori} probability, ultimately leading to the convergence in failure scenario recognition.

\subsection{Linear Quadratic Tracking for AUV}\label{section:LQT}

\begin{figure}
    \centering
    \includegraphics[width=0.7\columnwidth]{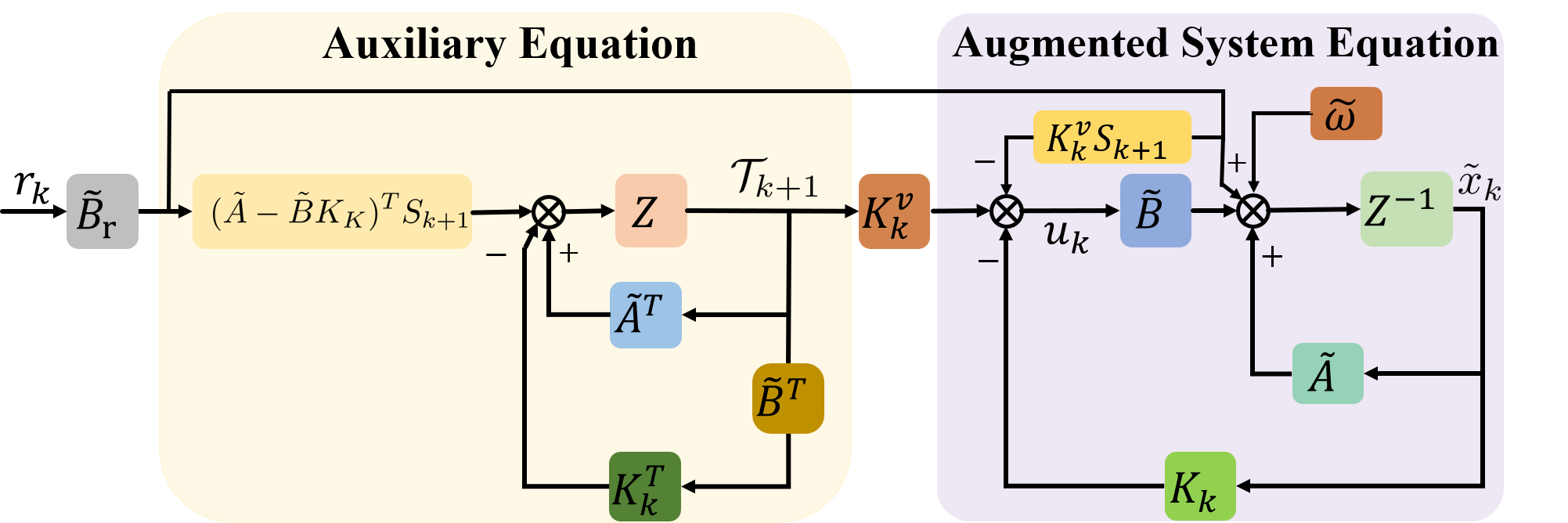  }
    \caption{Block diagram for LQT control for the AUV}
    \label{fig:17}
\end{figure}

This section derives linear quadratic tracking (LQT) control for AUV for a given failure scenario. 
We define the following output error for the LQT control
\begin{equation}
    \bm{e}_k=\bm{r}_k-\bm{y}_k,
\end{equation}
where $\boldsymbol{r}_k=[x_r(k),y_r(k),z_r(k),\phi_r(k),\varphi_r(k),\psi_r(k)]^T$ represent the trajectory to be followed at the time instant $k$, and $\bm{y}_k=[x(k),y(k),z(k),\phi(k),\varphi(k),\psi(k)]^T$ is the measured position and attitude vector of the AUV at the time instant $k$. We define the integral state $\bm{z}_k$ as
\begin{equation}
    \bm{z}_{k+1}=\bm{z}_k+\bm{e}_k.
\end{equation}
$\bm{z}_k$ represents the accumulated output error. Then we define the augmented state vector
\begin{equation}
    \tilde{\bm{x}}_k=\begin{bmatrix}
        \bm{x}_k\\\bm{z}_k
    \end{bmatrix},
\end{equation}
and the state space equation for AUV can be rewritten as:
\begin{equation}
\begin{aligned}
    \tilde{\bm{x}}_{k+1}&=\tilde{\bm{A}}_k\tilde{\bm{x}}_k+\tilde{\bm{B}}(k,\Gamma)\bm{u}_k+\tilde{\bm{B}_r}\bm{r}_k+\bm{N}\tilde{\omega}_k\\
    \bm{y}_k&=\tilde{\bm{H}}\bm{x}_k+v_k\\
    \Gamma &\in \Omega \triangleq\big \{\Gamma_1, \Gamma_2,\ldots,\Gamma_n\big \},
\end{aligned}
\end{equation}
where
\begin{equation}
\begin{aligned}
    \tilde{\bm{A}}(k)& =\begin{bmatrix}
       \bm{A}(k)&0\\-\bm{H}&\bm{I}
   \end{bmatrix},\tilde{\bm{B}}(k,\Gamma)=\begin{bmatrix}
       \bm{B}(k,\Gamma)\\0
   \end{bmatrix},
   \tilde{\bm{B}_r}=\begin{bmatrix}
       0\\\bm{I}
   \end{bmatrix},\\
   \tilde{\omega}(k)&=\begin{bmatrix}
      \omega_k\\v_k
   \end{bmatrix},\bm{N}=\begin{bmatrix}
      \bm{I}&0\\0&-\bm{I}
   \end{bmatrix},\tilde{\bm{H}}=\begin{bmatrix}
      \bm{H}&0
   \end{bmatrix},
\end{aligned}
  \nonumber
\end{equation}
where the observation matrix $\bm{H}$ is a full-rank matrix with certain sparsity the observation matrix of the system, and $\tilde{H}$ is the augmented matrix. In this way, we transform the cost function in equation\ref{Teq:cost_func} as  
\begin{equation}
    \tilde{J}=\frac{1}{2}\mathbb{E}\left[\tilde{\bm{x}}_{N}^T{P}\tilde{\bm{x}}_{N}+\sum^{N-1}_{k=0}(\tilde{\bm{x}}_{k}^T\tilde{Q}\tilde{\bm{x}}_{k}+\bm{u}_kR\bm{u}_k)\right],
    \label{cost fun}
\end{equation}
where $\tilde{Q}$ is the augmented state weight matrix. Since the reference trajectory $\bm{r}_k$ does not vary abruptly, $\tilde{Q}$ can be computed by
\begin{equation}
\tilde{Q}=\begin{bmatrix}\tilde{\bm{H}}^TQ\tilde{\bm{H}}&\tilde{\bm{H}}^TQ\\
Q\tilde{\bm{H}}&Q
\end{bmatrix}.
\nonumber
\end{equation}
The objective for LQT control is to determine the optimal control input $u^*_k$ that minimizes the cost function $J$. To this end, we apply the Maximum Principle to formulate a Hamiltonian function \cite{lewis2012optimal} $\mathcal{H}_k$:
\begin{equation}
    \mathcal{H}_k = \tilde{x}_k^T \tilde{Q} \tilde{x}_k + u_k^T R u_k + \tilde{\lambda}_{k+1}^T \left( \tilde{A} \tilde{x}_k + \tilde{B} u_k + \tilde{B}_r r_k \right),
\end{equation}
where $\lambda \in \mathbf{R}^n $ is a Lagrange multiplier~\cite{Murray2009}\cite{lewis2012optimal}. Then we have the AUV's state equation as
\begin{equation}
 \tilde{\bm{x}}_{k+1} =\left( \frac{\partial \mathcal{H}_k}{\partial \tilde{\lambda}_{k+1}}\right)^T=\tilde{\bm{A}} \tilde{\bm{x}}_k + \tilde{\bm{B}} u_k + \tilde{\bm{B}}_r r_k, \label{lqg:1}
\end{equation}
and the AUV's costate equation as
\begin{equation} \tilde{\lambda}_{k} =\left( \frac{\partial \mathcal{H}_k}{\partial \tilde{\bm{x}}_{k}}\right)^T=\tilde{Q} \tilde{\bm{x}}_{k}+\tilde{\bm{A}}^T\tilde{\lambda}_{k+1},\label{lqg2}
\end{equation}
and the Stationarity condition as
 \begin{equation}
 \begin{aligned}
  0&=\left( \frac{\partial \mathcal{H}_k}{\partial \bm{u}_{k}}\right)^T=R\bm{u}_k+\tilde{\bm{B}}^T\tilde{\lambda}_{k+1}\\
&\Rightarrow \bm{u}_k=R^{-1}\tilde{\bm{B}}^T\tilde{\lambda}_{k+1},\label{lqg3} 
 \end{aligned}
 \end{equation}
which are subject to the boundary conditions
\begin{equation}
\begin{aligned}
\tilde{\bm{x}}_0&=\begin{bmatrix}
    \bm{x}_0\\0
\end{bmatrix}
~~\textit{ and $\bm{x}_0$ is given},\\
\tilde{\lambda}_N&=\tilde{Q}\bm{x}_N.
\end{aligned}
\end{equation}
Note that in~\eqref{lqg:1}, the state space equation adds a reference trajectory term. This renders that the control law cannot be implemented, since the boundary conditions are split between times $k=0$ and  $k=N$. it is reasonable to assume that for all $k \leq N $:
 \begin{equation}
     \tilde{\lambda}_k=S_k\tilde{\bm{x}}_k-\mathcal{T}_k\label{lqg4}
 \end{equation}
for some yet unknown auxiliary sequences $S_k$ and $ \mathcal{T}_k$. This assumption holds if consistent equations can be found for $ S_k$ and $ \mathcal{T}_k$. Combined \eqref{lqg:1} - \eqref{lqg4}, we can gain the  recursions of auxiliary sequences $S_k$ and $\mathcal{T}_k$ as
 \begin{equation}
    S_k=\tilde{Q}+\tilde{\bm{A}}^TS_{k+1}(\tilde{\bm{A}}-\tilde{\bm{B}}K_k),\label{lqg5}
 \end{equation}
\begin{equation}
   \mathcal{T}_k=(\tilde{\bm{A}}-\tilde{\bm{B}}K_k)^T\mathcal{T}_{k+1}+(\tilde{\bm{A}}-\tilde{\bm{B}}K_k)^TS_{k+1}\tilde{\bm{B}}_rr_k, \label{lqg6}
\end{equation}
where
\begin{equation}
    \begin{aligned} K_K&=R+\tilde{\bm{B}}^TS_{k+1}\tilde{\bm{B}})^{-1}\tilde{\bm{B}}^TS_{k+1}\tilde{\bm{A}}\\
    K^v_K&=(R+\tilde{\bm{B}}^TS_{k+1}\tilde{\bm{B}})^{-1}\tilde{\bm{B}}^T\\
     \mathcal{T}_N&= (\tilde{\bm{A}}-\tilde{\bm{B}}K_k)^T\tilde{Q}\tilde{\bm{B}}_rr_N\\
      S_N&=\tilde{Q},
    \end{aligned}
\end{equation}
and we can obtain the optimal control law for the AUV failure model $\Gamma_s$ as
\begin{equation}
    \bm{u}_{k,s}=K^v_K\mathcal{T}_{k+1}-K_K\tilde{x}_k-K^v_KS_{k+1}\tilde{\bm{B}}_rr_k.\label{lqg7}
\end{equation}
We show the LQT control for AUV in \Cref{fig:17}. There are three critical terms in the control law derivation that (i) a \emph{feedback} term $\tilde{\bm{x}}_k$ with a gain coefficient determined by Riccati equation in \eqref{lqg5} (ii) a \emph{feedforward} term $\mathcal{T}_k$ with a gain derived from reference $r_k$ by the auxiliary difference equation in \eqref{lqg6} (iii) a \emph{feedback} term $r_k$ with a gain determined by the Riccati equation in \eqref{lqg5}.

However, the above derivation assumes full knowledge about the AUV state. This does not hold in real control systems and AUV state is not fully observable due to limited sensor measurements. This necessitate state estimation that is crucial for the control design. 

\subsection{AUV State Estimation}\label{EKF}
This section details the estimation for AUV state. We consider process noise and measurement noise AUV dynamics, and we adopt Extended Kalman Filter (EKF) in the estimation. We assume a single thruster failure under the condition $\varphi$. 
At the time instant $k$, the optimal state estimate $\hat{\tilde{\bm{x}}}(k|k,\varphi)$ and we use it to predict the \emph{prior} state estimate $\hat{\tilde{\bm{x}}}(k+1|k,\varphi)$ at time instant $k+1$. Then we observe the system output at $k+1$ based on which we can obtain the measurement $\bm{z}(k+1)$. We incorporate $\bm{z}(k+1)$ to update the \emph{prior} estimate $\hat{\tilde{\bm{x}}}(k+1|k,\varphi)$, resulting in the posterior estimate $\hat{\tilde{\bm{x}}}(k+1|k+1,\varphi)$, which represents the optimal estimate of the system state at time $k+1$. We summarize the iterative state estimation below:
\begin{equation}
\begin{aligned}
&\text{State Estimate Update}\\
&\hat{\tilde{\bm{x}}}(k+1|k+1, \varphi)=\hat{\tilde{\bm{x}}}(k+1|k, \varphi)+F(k+1, \varphi) \left[ z(k+1)-\right.\\
&\tilde{\bm{H}} (k+1, \varphi)\hat{\tilde{\bm{x}}}(k+1|k, \varphi) ] \\
&\text{State Estimate  Propagation}\\
&\hat{\tilde{\bm{x}}}(k+1|k, \varphi)=\tilde{\bm{A}}(k+1|k, \varphi)\hat{\tilde{\bm{x}}}(k|k, \varphi)+\tilde{\bm{B}}(k+1|k,\varphi)\bm{u}(k) \\
&\text{Kalman Gain Propagation}\\
&\ F(k+1, \varphi)=P(k+1|k, \varphi)\tilde{\bm{H}}\left(k+1, \varphi)\right[\tilde{\bm{H}}(k+1,\varphi)\\
&P(k+1|k,\varphi)\tilde{\bm{H}}^T(k+1,\varphi) 
+\bm{R}(k+1)]^{-1}\\
&\text{Error Covariance Propagation}\\
&P(k+1|k, \varphi)=\tilde{\bm{A}}(k+1|k, \varphi)P(k|k, \varphi)\tilde{\bm{A}}^{T}(k+1|k, \varphi)\\
&+\tilde{Q}(k+1|k, \varphi)\\
&\text{Error Covariance Update}\\
&P(k+1|k+1, \varphi)=\\ 
&\left[\bm{I}-F(k+1, \varphi)\tilde{\bm{H}}(k+1,\varphi)\right]P(k+1|k, \varphi)
\end{aligned}
\nonumber
\end{equation}
Note that the above estimation applies for any AUV thruster failure model in the space $\Omega$. Therefore, for each of the failure models, an optimal control law can be derived. Then the AUV can select from a bank of control laws the most appropriate one that fits the current AUV status best. However, traversing all the control laws at each time instant is unnecessary, and frequent switch between control laws due to shifted failure model can lead to jitter during switching that results in temporary loss of control and degraded system performance. Furthermore, the way that the control strategy switches from one to anther leads to ``hard switching'' that is computationally expensive and inefficient in the control.
This necessitates an adaptive control strategy that can coordinate the control laws for different failure models and facilitate the control in case of failure model shift. 
 
\subsection{Bayesian Probability for AUV Failure Models}\label{section:posterior_probability_update}
This section presents an adaptive approach in determining the AUV failure model via Bayesian method. We assume a bank of optimal control laws $U^*$ derived from individual failure models in the model space $\Omega$. 
To avoid hard switching between control laws for individual failure mdoels, we introduce a Bayesian posterior probability-based method to enable adaptive and ``soft switching'' between failure models and the corresponding control laws in  $U^*$, as elaborated below.

According to Lainiotis' partition theorem~\cite{LAINIOTIS_1971,Lainiotis1974}, the cost function defined in \eqref{cost fun} can be reformulated as a probabilistic integration over the model space $\Omega$. Given the observation sequence $\Upsilon_k\triangleq \{z(0),z(1),z(2),\cdots z(k)\}$ up to time instant $ k$, we define the global cost $\mathbb{J}$ as the Bayesian posterior-weighted expectation of the minimum cost for all candidates $\varphi \in \Omega$:
\begin{equation}
\begin{aligned}
    & \mathbb{J}=\mathbb{E}\bigg{\{}\underbrace{\text{MIN}_{}}_{u_k,\ldots,u_{N-1}}\frac{1}{2}\mathbb{E}\bigg{\{}\bigg{(}\bm{x}^T_NP\bm{x}_N+\\
    & \sum^{N-1}_{k=0}\big{(}\bm{x}_k^TQ\bm{x}_k+\bm{u}_kR\bm{u}_k\big{)}\bigg{)}\bigg{|}\varphi,\Upsilon_k\bigg{\}}\bigg{/}\Upsilon_k \bigg{\}},
\end{aligned}
\label{cost2}
\end{equation}
Where $\varphi$ represents the AUV condition that corresponds to a specific failure model in $\Omega$. The cost function in \eqref{cost2} can be rewritten as weighted sum of the control laws for each individual $\varphi$, weighted by the posterior probability of $p(\varphi)$. The posterior probability reflects the likelihood of each failure model being the actual model. The control objective turns to minimize the cost function in \eqref{cost2} based on the \emph{posteriori} probabilities under the condition $\varphi$.

According to Bayes' theorem~\cite{Lainiotis1974,Deshpande_1973}, the \emph{posteriori} probability $p(\varphi|\Lambda_k)$ given the system observations $\Lambda_k$ can be calculated by 
\begin{equation}
    p(\varphi|\Lambda_{k})=\frac{L(k|\varphi)}{\int_{\Omega_{\varphi}}L(k/\varphi)p(\varphi|\lambda_{k-1})d\varphi}p(\varphi|\Lambda_{k-1})
    \label{eq:24}
\end{equation}
\begin{equation}
\begin{aligned}
   & L(k|\varphi)= \left|P_{z}(k|k-1, \varphi)\right|^{-\frac{1}{2}}\cdot\exp\big[-\frac{1}{2}\|\\&\tilde{z}(k|k-1, \varphi)\|^{2}
  {P^{-1}_{z}}_{(k|k-1, \varphi)}\big] \\
  &  L(0|\varphi)=1
\end{aligned}
 \label{eq:23}
\end{equation}
\begin{equation}
\begin{aligned}
 &\tilde{z}(k|k-1,\varphi)\triangleq z(k)-\tilde{\boldsymbol{H}}(k,\varphi)\hat{x}(k|k-1,\varphi)\\
&P_{z}(k|k-1,\varphi)\triangleq \tilde{\boldsymbol{H}}(k,\varphi)P(k|k-1,\varphi)\tilde{\boldsymbol{H}}^{T}(k,\varphi) +\boldsymbol{R}(k) 
\end{aligned} 
\nonumber
\end{equation}
where $\boldsymbol{R}(k)$ is the covariance of the measurement noise $v(k)$, $L(.)$ is a likehood function, $P_{z}(k|k-1,\varphi)$ is the error covariance matrix, which can be calculated by EKF described in Section \ref{EKF}.

The computation for implementing the control is closely related to the admissible values of $\varphi$. 
In most control applications, the range of admissible values for $\varphi$ is continuous, which is generally computationally demanding. To address this issue, we in this work approximate the continuous parameter with a finite set of quantized points, so that the computation can be significantly reduced. 
We present the apriori probability for $\varphi$ as:
\begin{equation}
    p(\varphi)=\sum_{i=1}^{N}p(\varphi_{i}){\delta}(\varphi-\varphi_{i}),
\end{equation}
and we can rewrite \eqref{eq:24} as:
\begin{equation}
    p(\varphi_i|\Lambda_k)=\frac{L(k|\varphi_i)}{\sum_{j=1}^{\mathbf{M}}L(k|\varphi_j)p(\varphi_j|\Lambda_{k-1})}p(\varphi_i|\Lambda_{k-1}),
\label{eq:28}
\end{equation}
and the final control law $u^*(k)$ can be presented as
\begin{equation}
    u^*(k)=\sum\limits_{i=1}^{\mathrm{M}}\hat{u}(k|k,\varphi_i)p(\varphi_i|\Lambda_k).
    \label{eq:27}
\end{equation}

The calculations in \eqref{eq:24}-\eqref{eq:28} facilitates an adaptive process to identify and update AUV failure types based on the AUV's states. This iterative approach leads to the convergence of the estimated thruster failure to the correct one over the time using Bayesian inference. 
However, there is an issue with the convergence of failure identification that, if additional thrusters fail after an convergence, the already converged posterior probability of a failure might prevent the failure model shift. In that case, it looks like the thruster failure identification is locked due to converged posterior probability. Therefore, an unlocking strategy is necessary to ensure that the system can step out of the convergence and start failure type identification again.

\subsection{Lock-in and Unlock Strategy}\label{section:unlock_strategy}
This section details how we unlock failure identification and ensure proper failure model switching for the control. We look into the situation that at time $k$, the \emph{posterior} probability of the failure model has been locked, such that the \emph{posterior} probability for the failure model $i$ is $p(\varphi_i|\lambda_k) = 1$, while the \emph{posterior} probability of any another model $j$ ,for $j \in \Omega$ and $i \neq j$, equals 0.  In the following $m$ sampling periods after time $k$, the \emph{posterior} probability of the failure model $i$ evolves as follows:
\begin{equation}
   \begin{aligned}
p(\varphi_i|\lambda_{k+m})&=\frac{L(k+m|\varphi_i)}{\sum_{j=1}^{\mathbf{M}}L(k+m|\varphi_j)p(\varphi_j|\lambda_{k+m-1})}p(\varphi_i|\lambda_{k+m-1})\\
&=\frac{1}{1+\frac{p(\varphi_j|\lambda_{k+m})}{p(\varphi_i|\lambda_{k+m})}\sum_{j=1}^{\mathbf{M}}\prod_{n=1}^m\frac{L(k+n|\varphi_j)}{L(k+n|\varphi_i)}}\\
\label{eq:29}
\end{aligned}
\end{equation}
Equation \eqref{eq:29} reflects the updated \emph{posterior} probability when failure type shifts. The probability update incorporates the likelihood and the cumulative effect of failure model switching over $m$ time intervals~\cite{Kailath2000}. This enables dynamic adjustment of failure model selection, ensuring that the system control sustains during failure model transitions. Below we explain the derivation of the probability update.

We assume the following initial posterior probabilities for the failure $\varphi_i$ at time $k$\cite{Anderson2005}
 \begin{equation}
     \begin{aligned}
      p(\varphi_i|\lambda_k)=1-\varepsilon~~\\
p(\varphi_j|\lambda_k)=\frac{\varepsilon}{M},  
     \end{aligned}
 \end{equation}
where $M$ is the total number of failure models, $\varepsilon$ is a small constant that $0 < \varepsilon \ll 1$. The posterior probability at time instant $k+m$ can be rewritten as
 \begin{equation}
     p(\varphi_i|\lambda_{k+m})=\frac{1}{1+\frac{\varepsilon}{M(1-\varepsilon)}\sum_{j=1}^{\mathbf{M}}\prod_{n=1}^m\frac{L(k+n|\varphi_j)}{L(k+n|\varphi_i)}},
 \end{equation}
and we define $\xi_j$ as:
 \begin{equation}
   \xi_j=\sum_{j=1}^{\mathbf{M}}\prod_{n=1}^m\frac{L(k+n|\varphi_j)}{L(k+n|\varphi_i)}.
 \end{equation}
If the failure model corresponding $\varphi_j$ is not the actual failure type for the AUV, then this failure model is not able to generate observation data that sustain the identification of the $j$-th failure model over time. 
The likelihood function $L(k+n|\varphi_j)$ for will be much lower than that of the actual failure model $L(k+n|\varphi_i)$. As a result, the accumulated ratio
 \begin{equation}
     \prod_{n=1}^m \frac{L(k+n|\varphi_j)}{L(k+n|\varphi_i)}
 \end{equation}
will decrease exponentially. Furthermore, the predictions made by the $j$-th failure model will deviate from the actual observations, leading to an increasing residual $\tilde{z}$. Consequently, the cumulative sum  $\xi_j(k+m)$ will approach zero as $m$ increases~\cite{Tresp2005}.



From the above, it follows that for the $i$-th failure model (the actual failure model), the \emph{posterior} probability will converge to 1:
\begin{equation}
    p(\varphi_i|\lambda_{k+m})\rightarrow 1,~~~as~~~ m\rightarrow \infty
\end{equation}
Similarly, for the $j$-th model that beyond the actual model, we have $\xi_j(k+m) \to \infty$ and the posterior probability of the true model $i$ will approach 0
\begin{equation}
    p(\varphi_i|\lambda_{k+m})\rightarrow 0
\end{equation}
With the above, In other words, after time $k$, regardless of whether the system mode changes, the \emph{posterior} probability for the actual failure type will always approach 1 after $k$ time steps. 

\section{Simulations and Results}~\label{sec-simulation}

This section presents two simulations in to validating the developed fault-tolerant control for AUV in trajectory tracking. The first simulation considers AUV trajectory tracking with failures in two thrusters. The second one shifts the AUV from normal to failure condition and check how the control works when there is a model switching. 

\subsection{Simulation Settings}

\begin{figure}[htb]
    \centering
    \includegraphics[width=0.4\columnwidth]{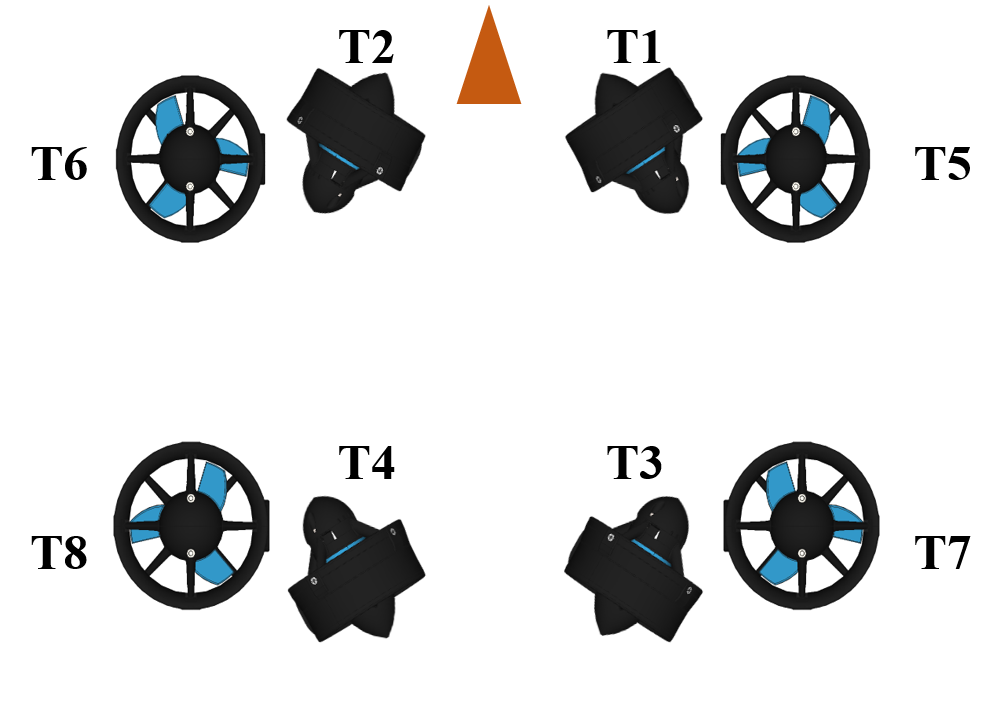}
    \caption{Thrusters on a BlueROV Heavy AUV.}
    \label{fig:4}
\end{figure}

In this simulation, an AUV model is developed to accurately represent the physical characteristics of a BlueROV2 Heavy, a commercial AUV produced by Blue Robotics~\cite{Amer_2023,Chaffre_2024}. We model the AUV with four vertical thrusters and four horizontal thrusters as shown in \Cref{fig:4}. The horizontal thrusters provide control over pitch, roll, and yaw of the AUV, while the vertical thrusters control over heave, pitch, and roll. The main physical and fluid dynamic parameters for the BlueROV2 Heavy model~\cite{Wu2018,jmse10121898} are presented in Table \ref{tab:param2}. We assume in the simulation that diagonally paired thrusters T5 and T8, as well as T6 and T7, operate synchronously in generating thrust in aligned directions. We conduct the simulations via MATLAB 2023b. 

\begin{table}[htb]
    \centering
    \caption{BlueROV2 dimensions and hydrodynamic parameters utilised in the simulations; data based on\cite{Wu2018},\cite{Walker_2021},\cite{Pettinelli2022}}
    \setlength{\tabcolsep}{3pt}
    \begin{tabular}{|c|c|c|c|}
    \hline
         Parameter  &  Nomenclature  & Value &  Unite\\ \hline
        Density of Seawater         & {$\rho$} & 1025 &{$kg/m^3$ }\\
        Dry Mass                    &{$m_d$}  & 11.5& {$kg$}\\
        Added Mass              & {$K_{\dot{p}}=M_{\dot{q}}=N_{\dot{r}}$} & -0.12 & {$kgm^2/rad$}\\
        Rotational Inertia, {$q$}     & {$\boldsymbol{I}_x=\boldsymbol{I}_y=\boldsymbol{I}_z$}  & 0.16 &{$kgm^2$} \\
        Added Mass ($surge$ )             &{$X_{\dot{u}}$}  & -5.5 &{$kg$} \\
        Added Mass ($sway$ )           &{$Y_{\dot{\nu}}$ }  & -12.7 &{$kg$} \\
        Added Mass ($heavy$ )          & {$Z_{\dot{\omega}}$}  & -14.57 &{$kg$} \\
        Restoring Moment Arm,$q$    & {$\bar{BG_z}$}  & 0.02 &{$m$} \\
        Maximum Output Thrust,      & {$T_{max}$}  & 40 & {$N$}\\
        Thruster Angle (Horizontal)        & {$\alpha_1,\alpha_2,\alpha_3,\alpha_4 $} & {$45^{\circ}$} & NULL\\
\hline
    \end{tabular}
    \label{tab:param2}
\end{table}

\subsection{Trajectory Tracking with Failure in AUV Thrusters}\label{section:4b}
This simulation assumes two of the horizontal thrusters are broken with $100\%$ thruster power lost. Following the AUV model in \Cref{fig:4}, we define the thruster failure space by 
 \begin{equation}
     \boldsymbol{\Omega}_{\varphi}=\{\boldsymbol{T}_1 \boldsymbol{T}_2,\boldsymbol{T}_1 \boldsymbol{T}_3,\boldsymbol{T}_1 \boldsymbol{T}_4,\boldsymbol{T}_2 \boldsymbol{T}_3,\boldsymbol{T}_2 \boldsymbol{T}_4,\boldsymbol{T}_3 \boldsymbol{T}_4 \}
     \label{eq:eq1}
 \end{equation}
We set the the noise and initial AUV conditions as
 \begin{equation}
 \begin{aligned}
     \boldsymbol{\nu}(t) &\sim (0, 0.02)\\
     \boldsymbol{\omega}(t) &\sim (0, 0.02)\\
     \boldsymbol{x}(0)&=\boldsymbol{0}\\
     p_1(0)&= p_2(0)= p_3(0)= p_4(0)=0.2\\
      p_5(0)&= p_6(0)=0.1,
 \end{aligned}
 \nonumber
 \end{equation}
where $p_i(0)$ ($i=1,2,\cdots, 6$) represents the initial \emph{posteriori} probability of the thruster failure models. We set a helical trajectory for the AUV tracking with the starting point as $(0,0,0)$, and we define the reference trajectory as
 \begin{equation}
     \begin{cases}
         x_d=sin(0.02t)\\
         y_d=cos(0.02t)\\
         z_d=0.02t
     \end{cases}
     \nonumber
 \end{equation}
 
This simulation separately models all the cases where two thruster faults in the parameter space $\Omega_{\varphi}$. A probability $p(t)$ is assigned to each of the failure models in $\Omega_{\varphi}$ by the proposed fault-tolerant control. During trajectory tracking, the thruster failure is identified while the AUV operates. 

\begin{figure}[htb]
    \centering
    \includegraphics[width=0.4\linewidth]{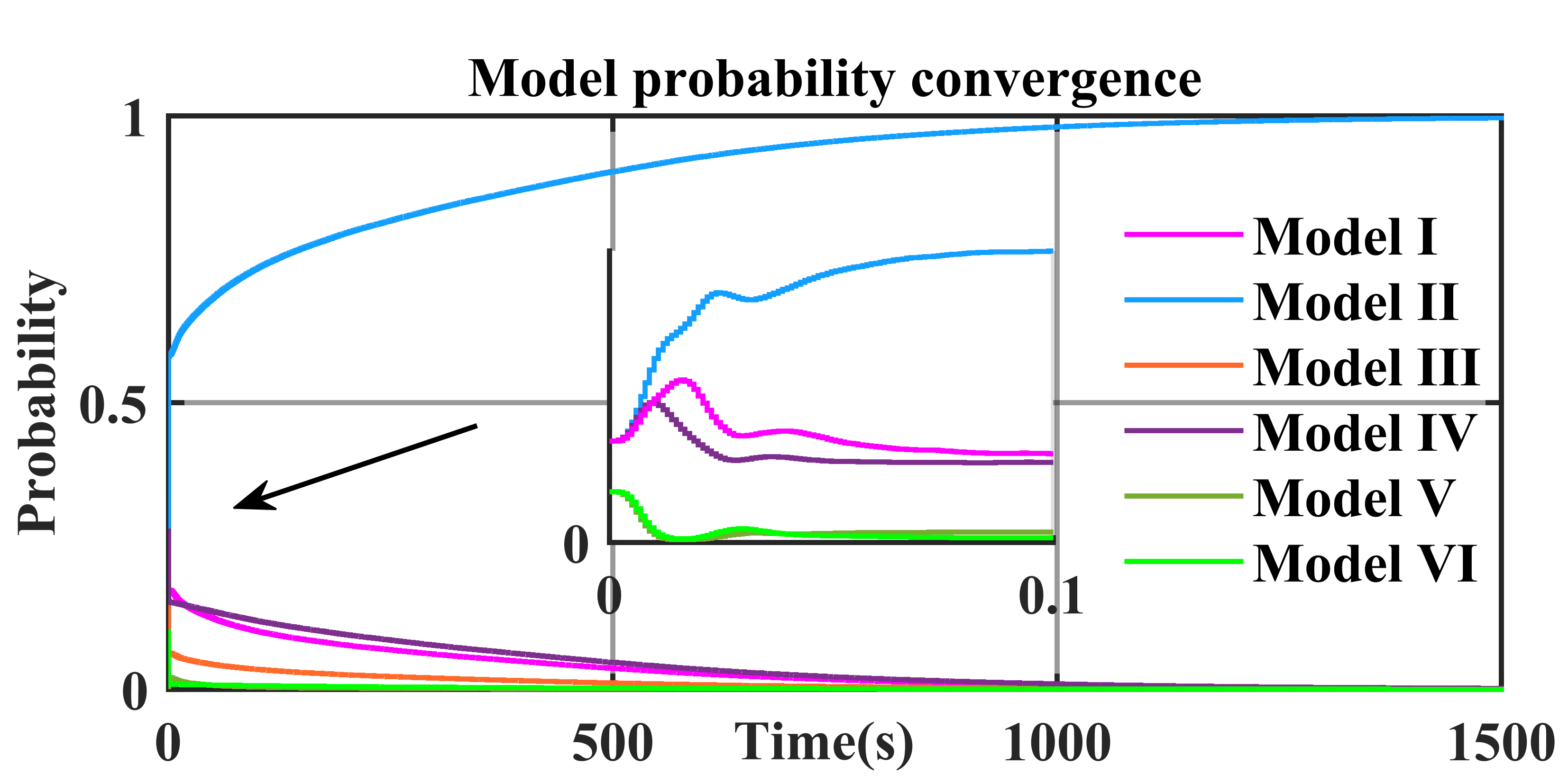}
    \caption{Convergence of different failure models.}
    \label{fig:5}
\end{figure}

This simulation considers the failures in thruster $\boldsymbol{T}_1$ and $\boldsymbol{T}_3$, corresponding to $\varphi=2$ in the $\Omega_{\varphi}$ model space. \Cref{fig:5} show the convergence of the \emph{posteriori} probability for each thruster failure models. The figure indicates that the \emph{posteriori} probability initially changes rapidly within the first 0.05s, followed by a gradual slowdown toward convergence. The probability for failure Model II converges to 1, while the probabilities of the others converge to 0. The result shows that AUV failure is correctly identified within 0.1s. Due to the different levels of similarity between the other five failure models and the actual model, their convergence decreases with different speed after the first 0.05s. 
\begin{figure}[htb]
\centering
\includegraphics[width=0.3\linewidth]{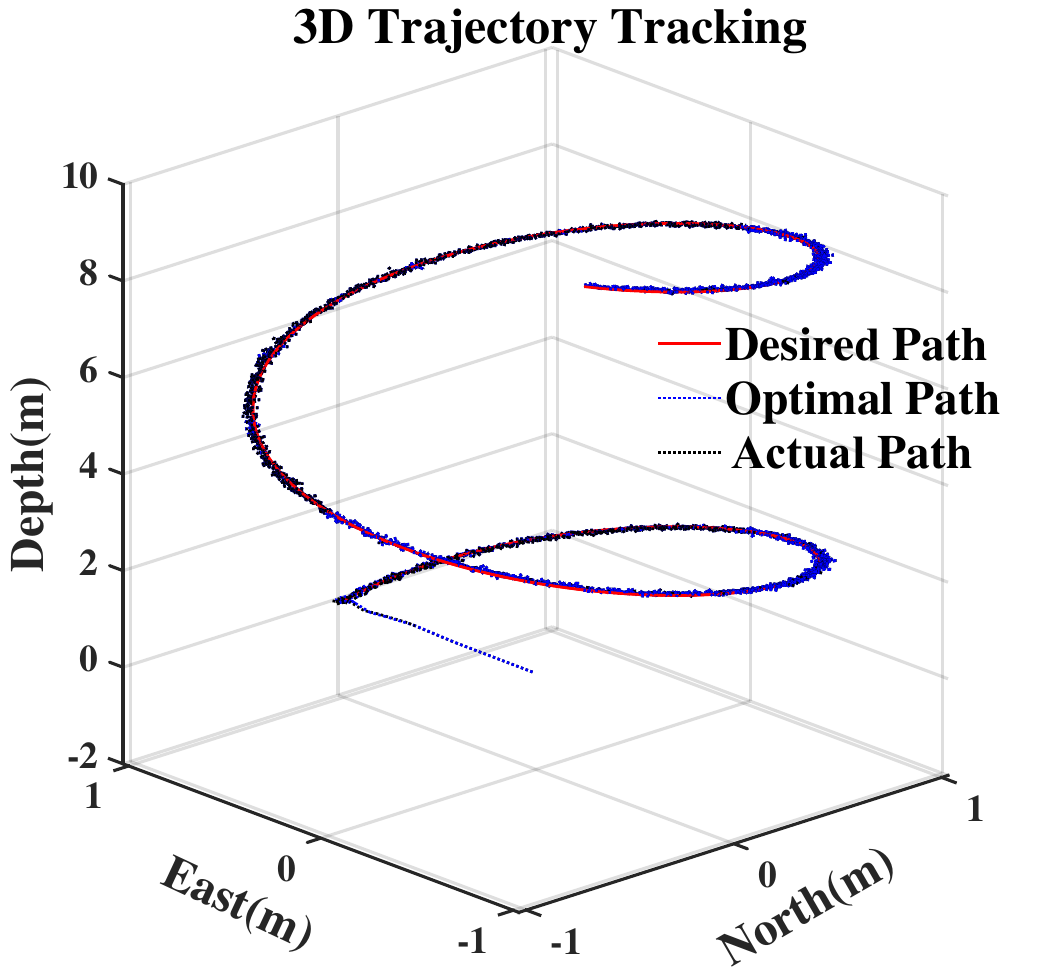}
\caption{Trajectory tracking results of different model and algorithm. ``Desired Path'' and ``Actual Path'' in mean the reference trajectory and the trajectory under thruster failures, respectively. ``Optimal Path'' means the AUV trajectory free from thruster failures.}\label{fig:6-7} 
\end{figure}

\begin{figure}[htb]
\centering
\begin{subfigure}{0.4\linewidth}
    \centering
    \includegraphics[width=\linewidth]{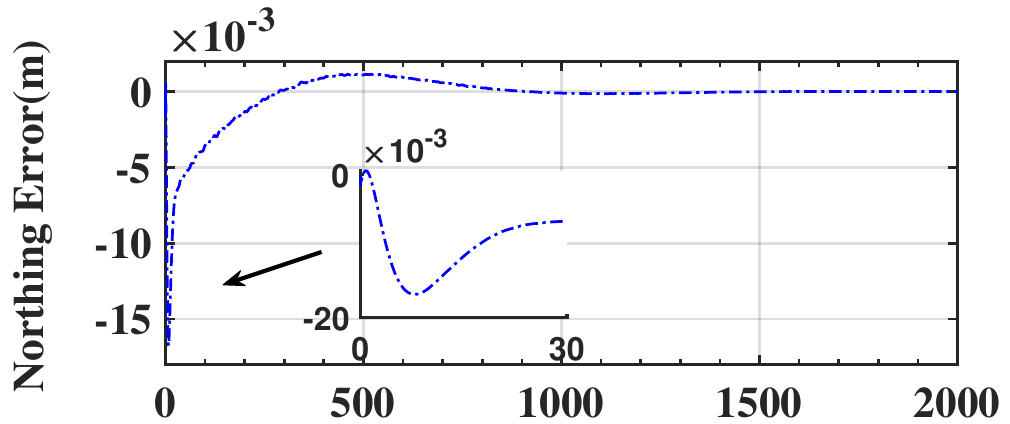 }
\end{subfigure}
\begin{subfigure}{0.4\linewidth}
    \centering
\includegraphics[width=\linewidth]{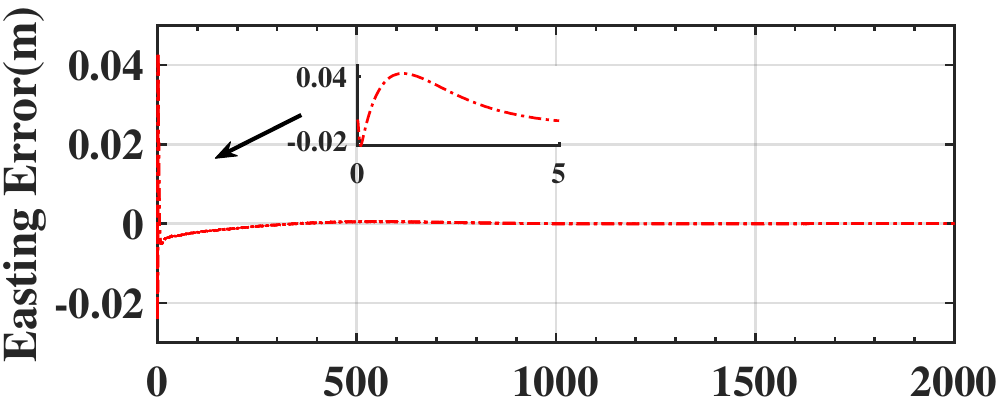 }
\end{subfigure} 
\begin{subfigure}{0.4\linewidth}
    \centering
\includegraphics[width=\linewidth]{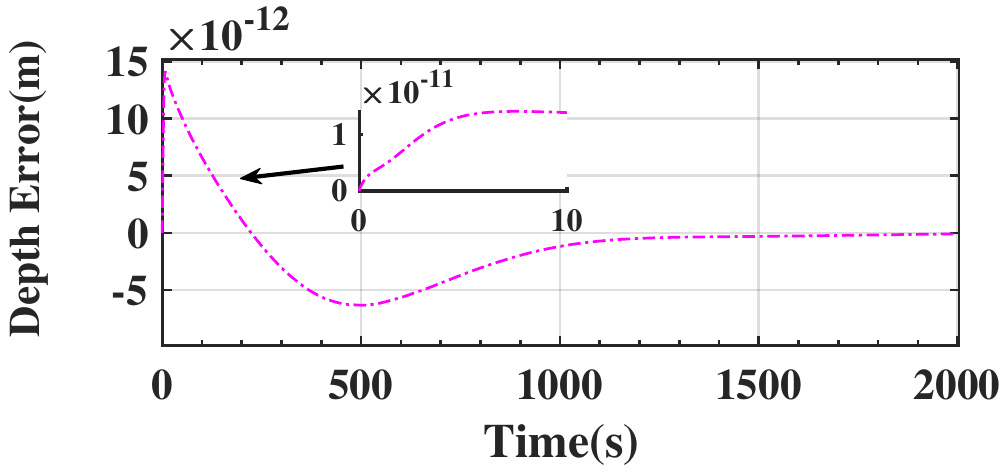 }
\end{subfigure} 
\caption{Error for the AUV trajectory tracking under the proposed fault-tolerant control.}
 \label{fig:8}
\end{figure}

Note that the trajectory tracking by the proposed fault-tolerant control is not affected by the failure identification. While we observe from \Cref{fig:6-7} that when the thrusters fail, the AUV experiences a reduced control. We quantify this reduced control by error metrics along the three axes shown in \Cref{fig:8}. The three axes indicate the the divergency of the AUV trajectory in the direction of north, east, and depth. The error in the direction of north shows that after a adjustment period of 30s, the AUV tracking achieves millimeter-level accuracy. The error in the direction of east is larger than the north error since the failed thrusters $\boldsymbol{T}_1$ and $\boldsymbol{T}_3$ are on the right side of the vehicle, which directly impacts AUV's horizontal motion. The error in the direction of depth is the slightest since the four vertical thrusters are functioning normally, leading to minor impact in the vertical motion of AUV.

To check how each thruster reacts during the fault-tolerant control, we present the control inputs from all the eights thruster across all the six thruster failure models shown in \Cref{eq:eq1}. We also compare with the control inputs by optimal control where the actual failure information is known to the controller. 

\begin{figure}[htb]
 \centering
\begin{subfigure}{0.45\linewidth}
    \centering
    \includegraphics[width=1\linewidth]{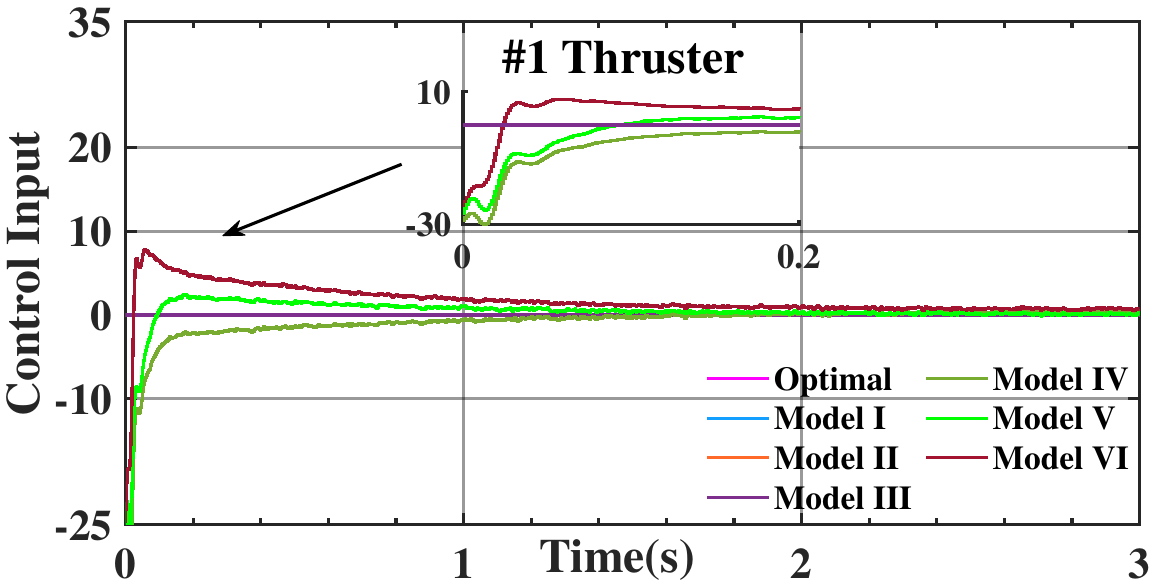}
    \label{fig:9}
\end{subfigure}\hfill
\begin{subfigure}{0.45\linewidth}
    \centering
\includegraphics[width=1\linewidth]{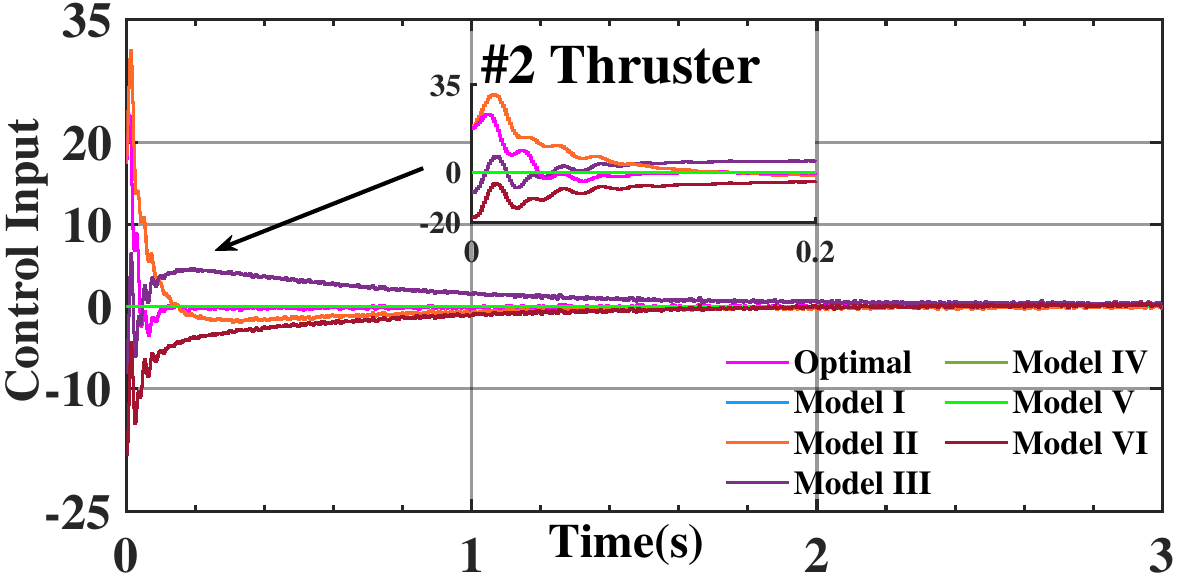}
    \label{fig:10}\
\end{subfigure}\hfill 
  \begin{subfigure}{0.45\linewidth}
    \centering
    \includegraphics[width=1\linewidth]{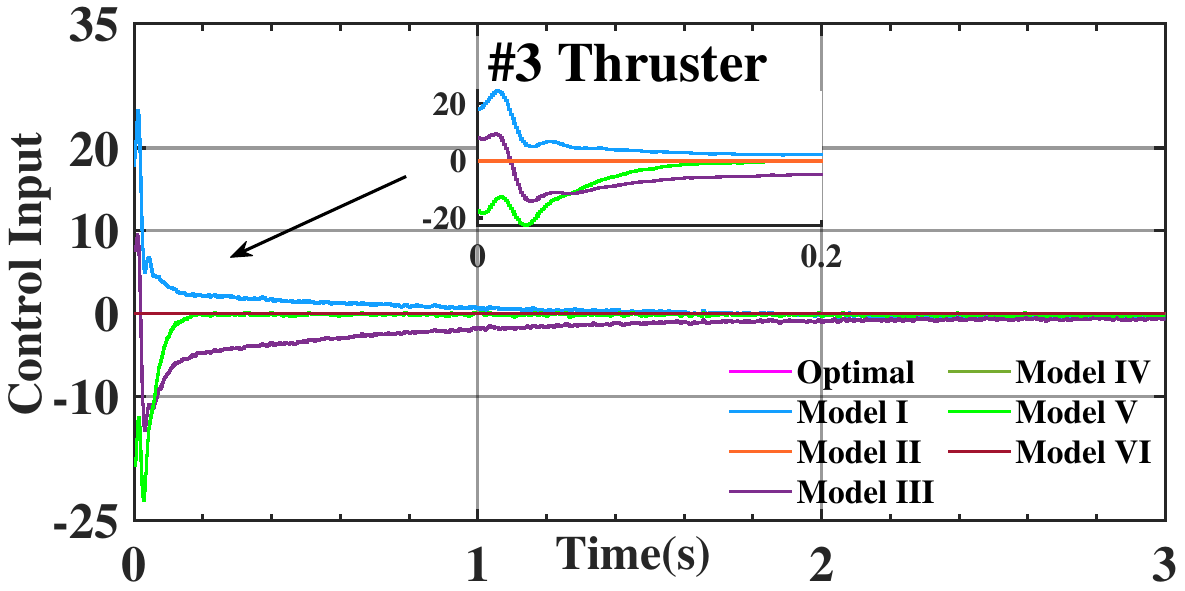}
    \label{fig:11}
\end{subfigure}\hfill
\begin{subfigure}{0.45\linewidth}
    \centering
\includegraphics[width=1\linewidth]{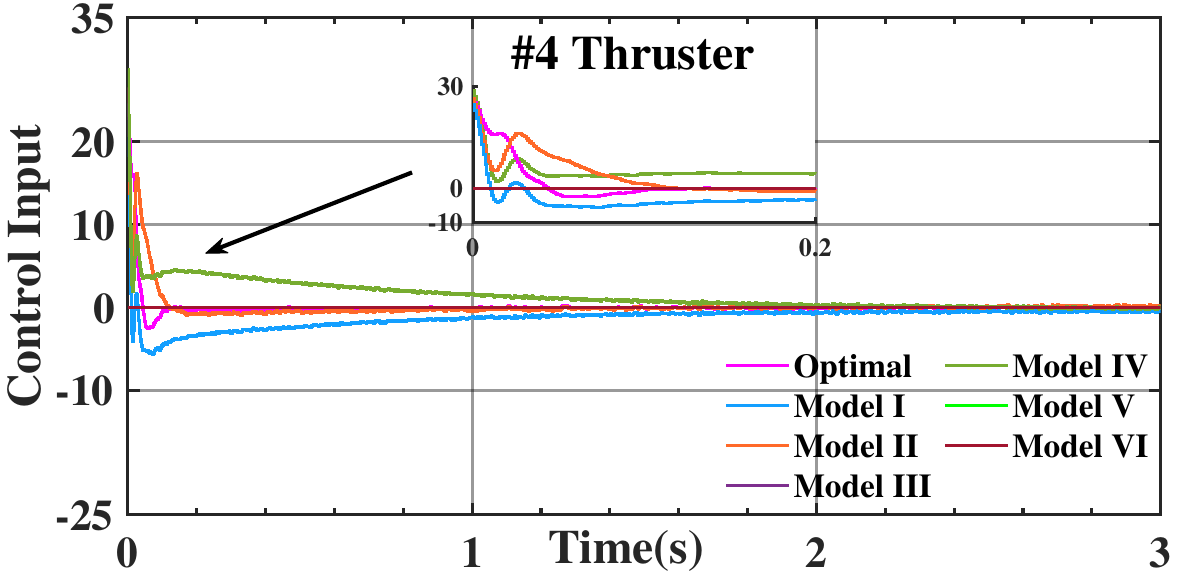}
    \label{fig:12}
\end{subfigure} 
      \caption{Comparison of control inputs by different failure models for horizontally arranged thrusters}\label{fig:9-12}
\end{figure}

The control inputs from the four horizontally arranged thrusters are shown in \Cref{fig:9-12}. $\boldsymbol{T}_1$ is involved in the failure model Model I, Model II, and Model III, and the control input for $\boldsymbol{T}_1$ is zero consistently. We observe the same pattern for $\boldsymbol{T}_2$ under various failure models. The failure models involving $\boldsymbol{T}_2$ include Model I, Model IV, and Model V, which results in control input of zero for $\boldsymbol{T}_2$. Compared with Model II, Model III, and Model VI, it is evident that Model II most approximates the optimal control, which aligns with the results shown in \Cref{fig:5}.

$\boldsymbol{T}_4$ is involved in failure models Model III, Model V, and Model VI. 
In the beginning of the control, the AUV horizontal movement manily relies on  $\boldsymbol{T}_4$ and only failure Model I and Model IV can achieving near-optimal control. As the AUV approaches the reference trajectory, $\boldsymbol{T}_2$ becomes increasingly critical. When Model I and Model IV can no longer maintain trajectory tracking, the failure model transfers to Model II. After a short period of adjustment, Model II successfully manage to stabilize the trajectory tracking.


\begin{figure}[htb]
 \centering
\begin{subfigure}{0.45\linewidth}
    \centering
    \includegraphics[width=1\linewidth]{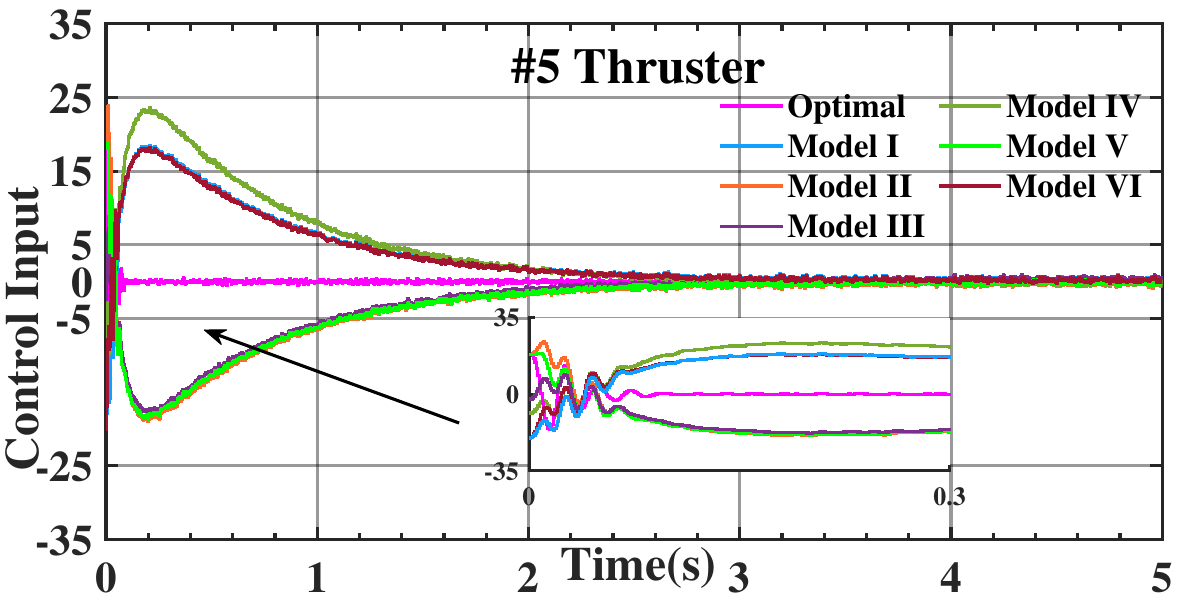}
    \label{fig:13}
\end{subfigure}\hfill
\begin{subfigure}{0.45\linewidth}
    \centering
\includegraphics[width=1\linewidth]{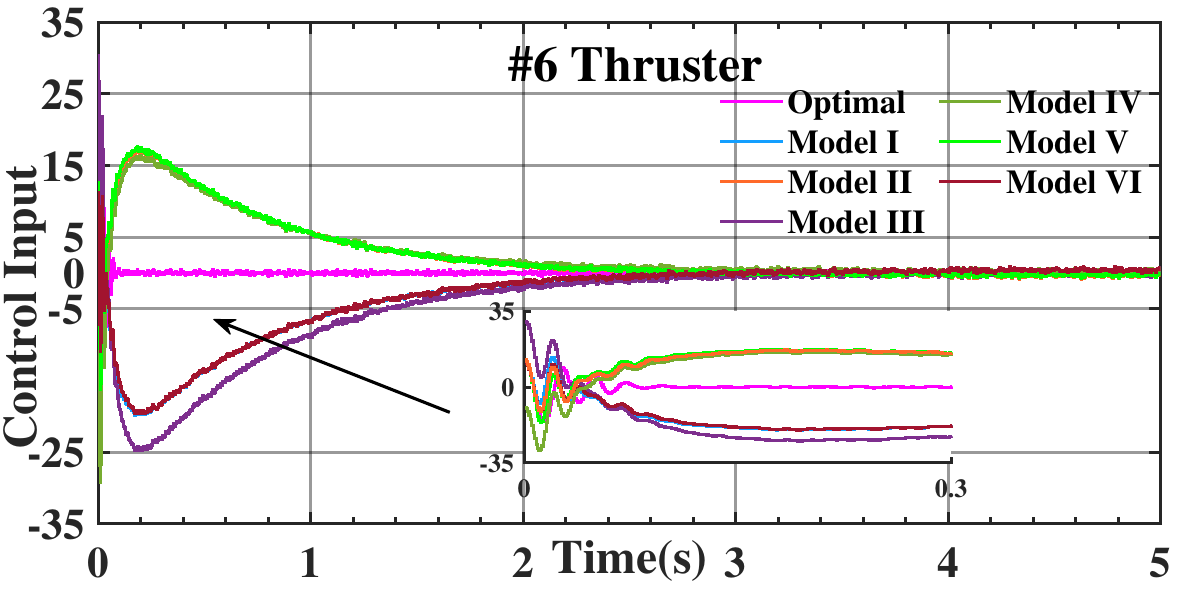}
    \label{fig:14}
\end{subfigure}   
  \begin{subfigure}{0.45\linewidth}
    \centering
    \includegraphics[width=1\linewidth]{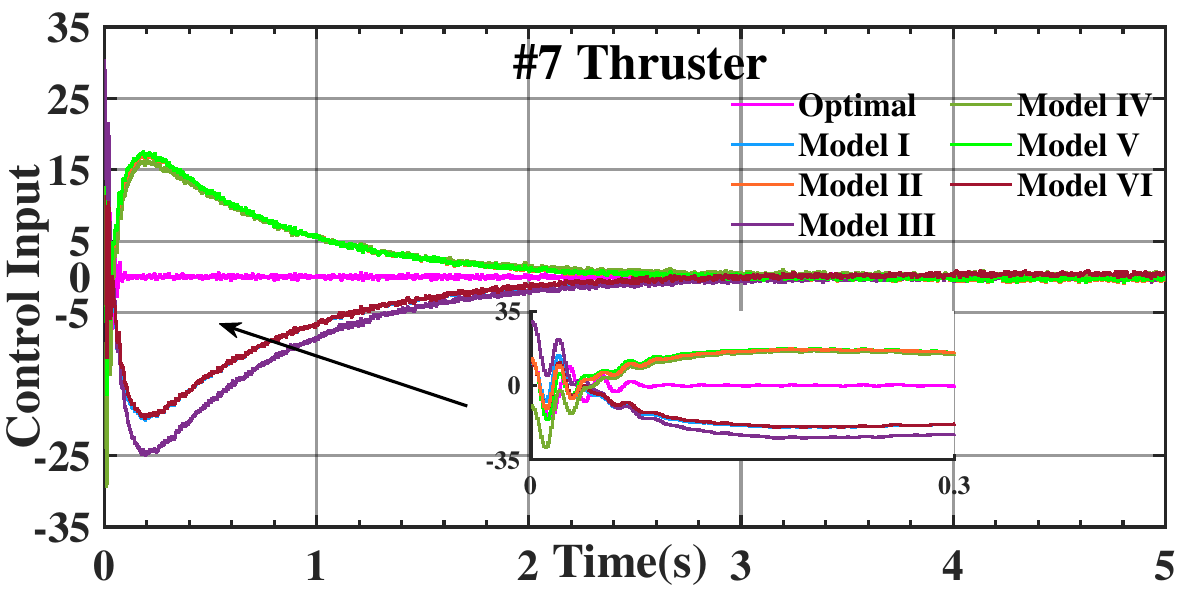}
    \label{fig:15}
\end{subfigure}\hfill
\begin{subfigure}{0.45\linewidth}
    \centering
\includegraphics[width=1\linewidth]{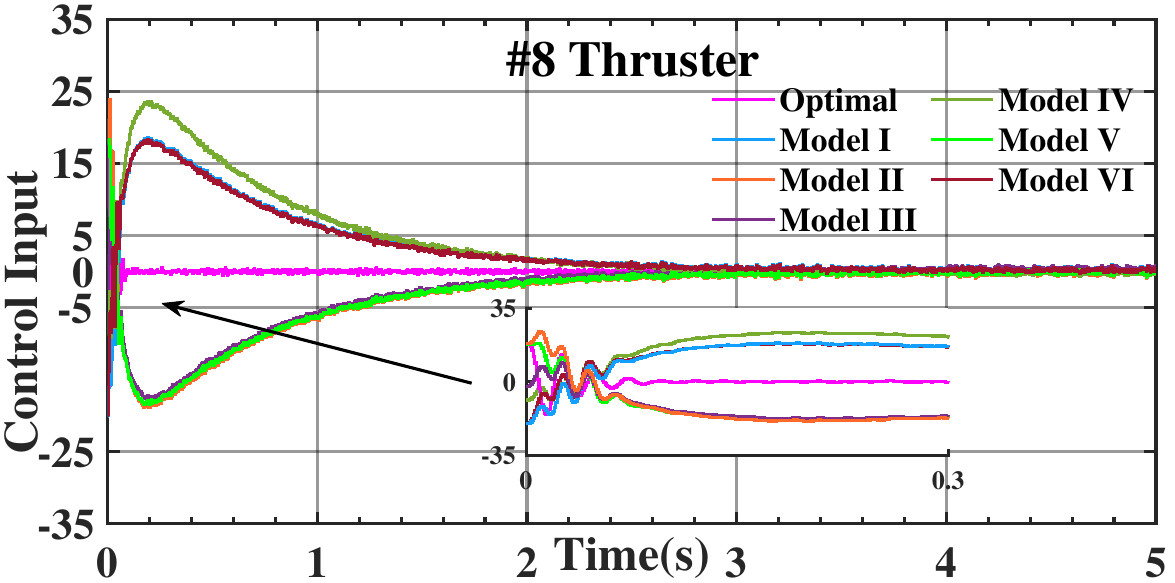}
    \label{fig:16}
\end{subfigure} 
      \caption{Comparison of control inputs for various models of a single thruster arranged vertically}\label{fig:13-16}
\end{figure}

The four vertically oriented thrusters play a crucial role in trajectory tracking. As illustrated in \Cref{fig:4}, thruster $\boldsymbol{T}_5$ and $\boldsymbol{T}_8$ consistently rotate in the same direction, and their thrust magnitudes are equal but in the opposite direction. The same applies to thrusters $\boldsymbol{T}_6$ and $\boldsymbol{T}_7$. We observe from \Cref{fig:13-16} that the the thrust under all the six failure models converge within the first 2.5s. The control input for Model II indicates a thrust of 20 for $\boldsymbol{T}_5$ and $\boldsymbol{T}_8$, and 16 for $\boldsymbol{T}_6$ and $\boldsymbol{T}_7$, resulting in an ascending motion of the AUV when thruster failure takes place.

Overall, in the case of thruster failure, the proposed strategy reconfigures the controller to maintain the control. The simulation results demonstrate that the proposed adaptive fault-tolerant control method can effectively track a trajectory in a three dimensional space even though two of the thrusters fail.

\subsection{Tracking Control with Shift in Thruster Failure Type}

This section simulates the proposed fault-tolerant control in the case of thruster failure type shift during the control. 
We keep all the simulation conditions the same with \Cref{section:4b}, except how the thruster failure takes place. 

\begin{figure}
    \centering
    \includegraphics[width=0.45\linewidth]{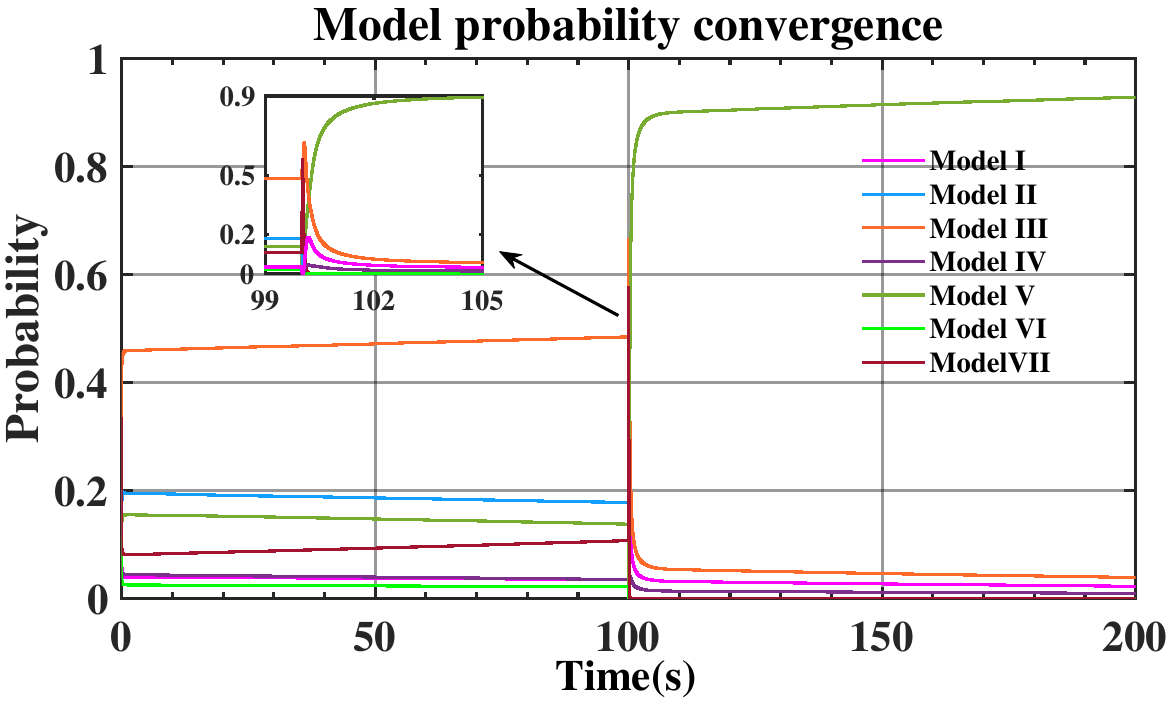}
    \caption{The probability convergence graph before and after model switching }
    \label{fig:20}
\end{figure}

This simulation initializes in a thruster failure where thrusters 1 and 4 are broken (failure Model III as shown in \Cref{fig:20}) and then switches to failure Model V (where thrusters 2 and 3 are inactive) at $t=100s$. \Cref{fig:20} shows that when failure type switches, the Bayesian probability of Model V increases quickly and reaches $p=0.9$ within 5 seconds, with other failure models decrease nearly to zero.

\begin{figure}[htb]
\centering
\begin{subfigure}{0.45\linewidth}
    \includegraphics[width=1\linewidth]{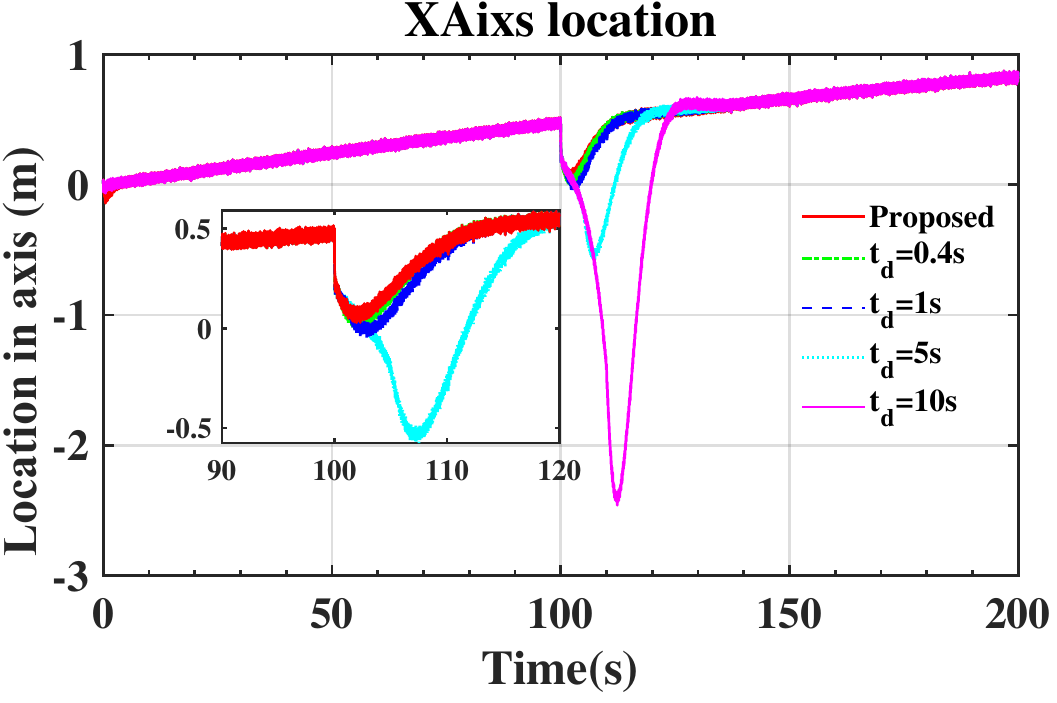}
    \label{fig:21}
\end{subfigure}\vfill
\begin{subfigure}{0.45\linewidth}
    \includegraphics[width=1\linewidth]{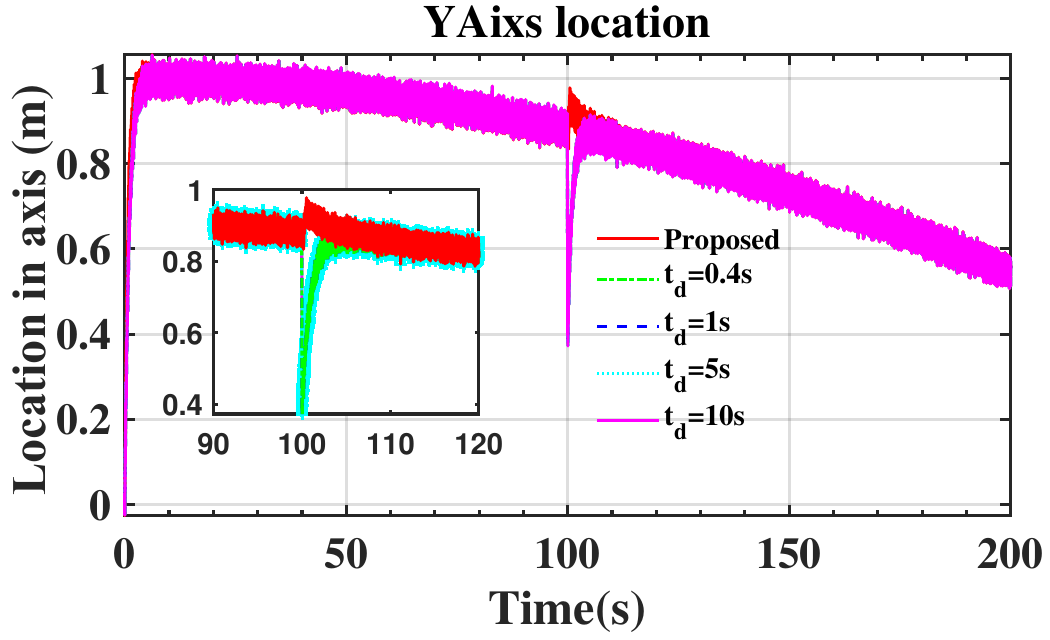}
\end{subfigure}
\caption{Comparison between the proposed approach and hard-switching in fault tolerant control.}\label{fig:22}
\end{figure}
\Cref{fig:22} compares the fault tolerant control between the proposed approach and hard-switching of fault detection with different detection intervals of 0.1, 1, 5, and 10 seconds. Note that at 100s, thruster 2 and 4 fail. Since thruster 2 and 4 only affect horizontal motion of the AUV, we only show the AUV trajectory in x and y axis in the results.

\Cref{fig:22} shows that for the hard-switching approach, the AUV motion is affected more severely as the detection interval increases. This is because the control scheme is not switched to the correct one until the failure is detected in the case of failure type shift. In comparison, the proposed fault tolerant control maintains the control with minimum ``pulse'' when the failure type shifts, manifesting the advantage of adaptive and soft-switch in fault tolerant control.


\begin{figure}[htb]
 \centering
\begin{subfigure}{0.5\linewidth}
    \centering
    \includegraphics[width=1\linewidth]{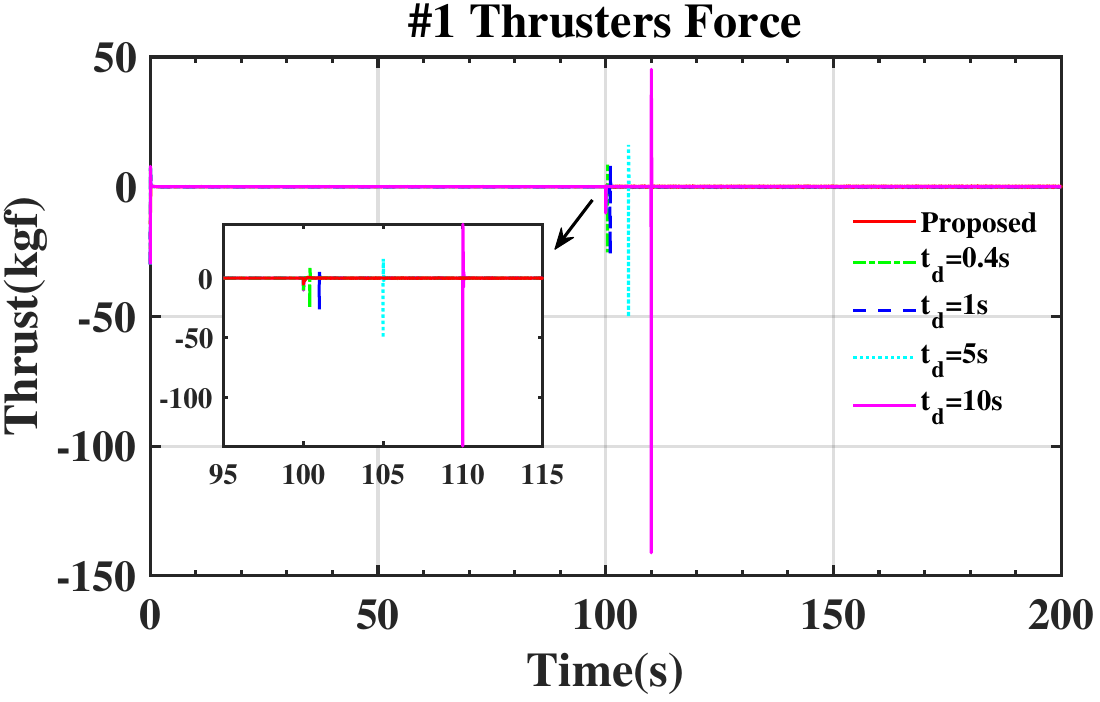}
    \label{fig:23}
\end{subfigure}\hfill
\begin{subfigure}{0.5\linewidth}
    \centering
    \includegraphics[width=1\linewidth]{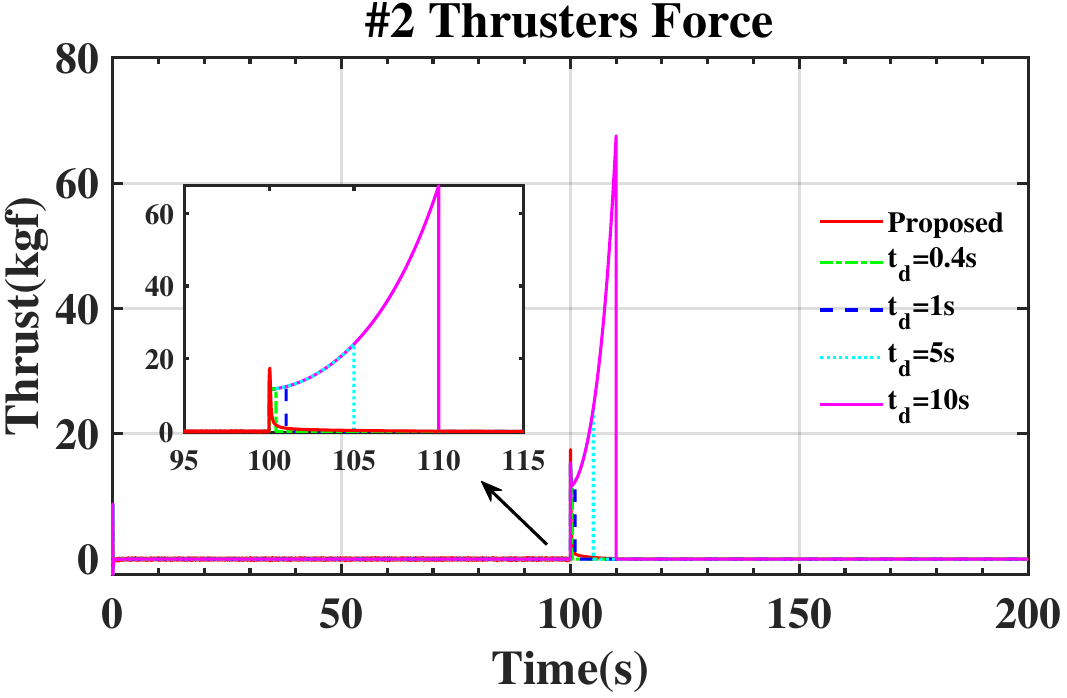}
    \label{fig:24}
\end{subfigure}\hfill
\begin{subfigure}{0.5\linewidth}
    \centering
    \includegraphics[width=1\linewidth]{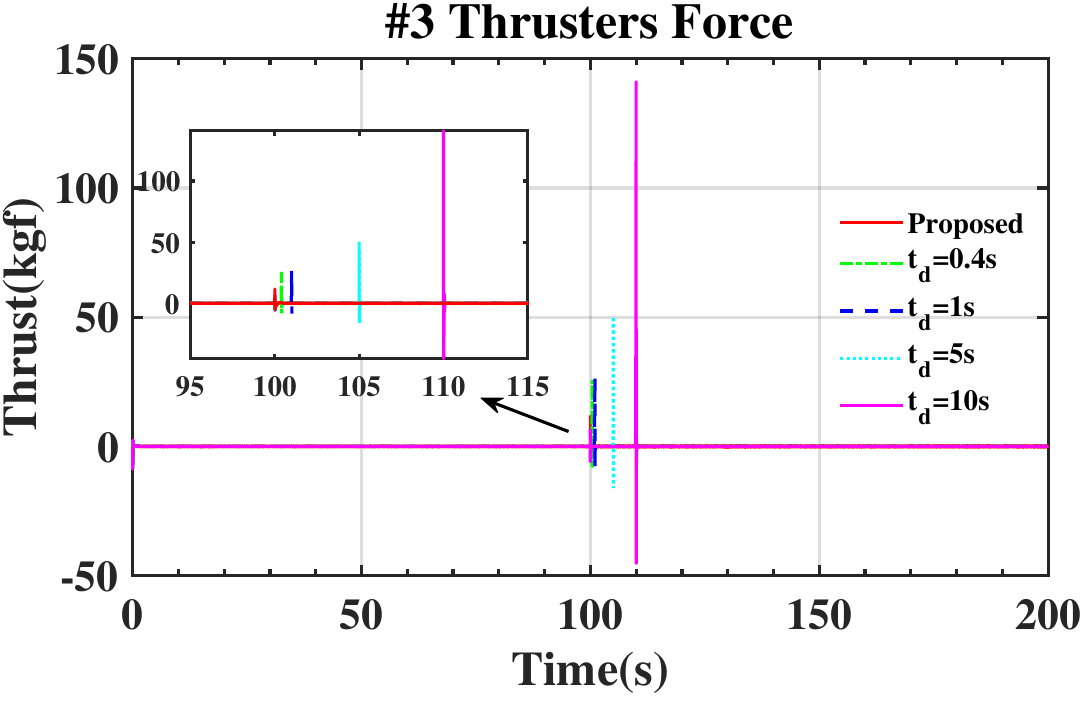}
    \label{fig:25}
\end{subfigure}\hfill
\begin{subfigure}{0.5\linewidth}
    \centering
    \includegraphics[width=1\linewidth]{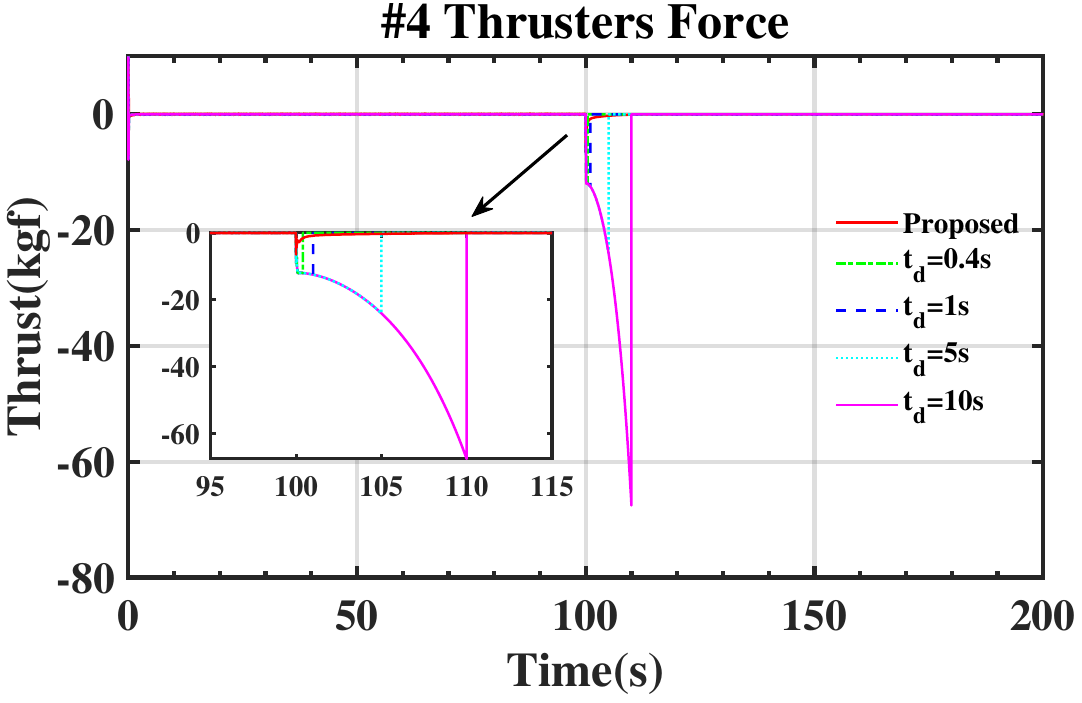}
\end{subfigure}
\caption{Comparison of thruster force from horizontally arranged thrusters between the proposed fault tolerant control and hard-switch approach with different fault detection intervals of 0.1, 1, 5, and 10 seconds.}\label{fig:26}
\end{figure}

\Cref{fig:26} shows how the control input varies when the thruster failure type shifts. We observe that at 100s, Thruster 1 experiences a peak adjustment of -5.02 kgf, while Thruster 3 undergoes adjustments of 11.65 kgf and -4.59 kgf before it stabilizes in maintaining the AUV trajectory tracking. Similarly, Thrusters 2 and 4 experience initial adjustments of 17.45 kgf and -6.48 kgf, respectively, and their thrust quickly turns to 0 due to thruster failure. \Cref{fig:26} indicates that hard-switching control results in meaningless adjustments in thrust force in Thrusters 2 and 4, despite their inoperative state. In comparison, the proposed adaptive fault-tolerant control method yields the minimum adjustment in thruster force. 

\begin{figure}[htb]
 \centering
\begin{subfigure}{0.5\linewidth}
    \centering
    \includegraphics[width=1\linewidth]{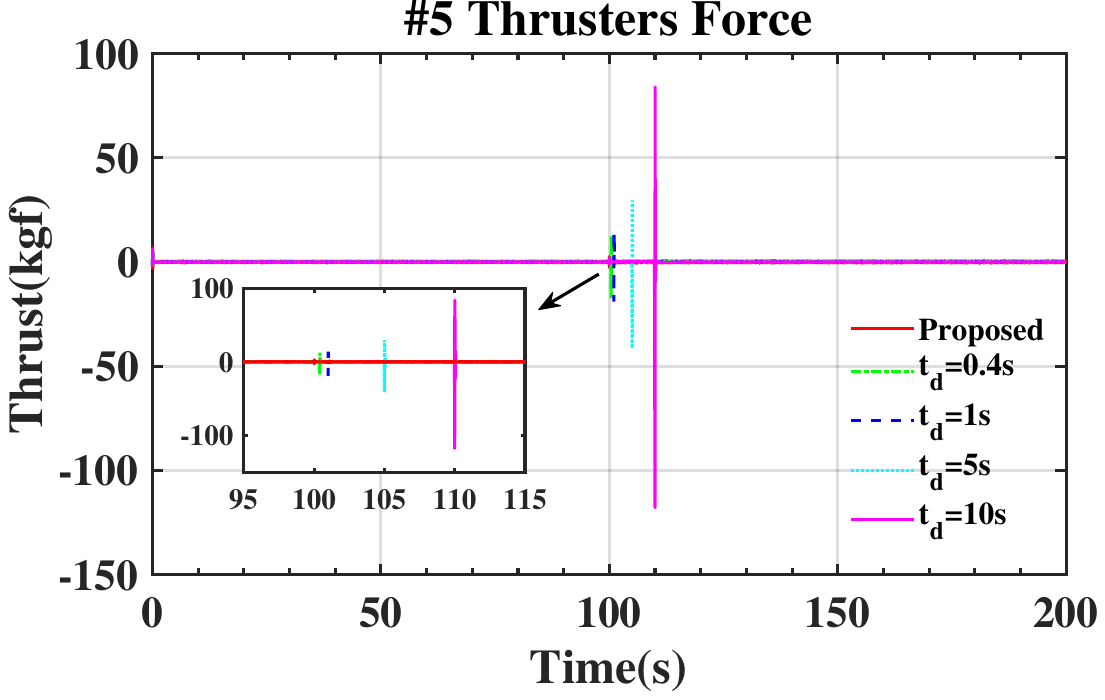}
    \label{fig:27}
\end{subfigure}\hfill
\begin{subfigure}{0.5\linewidth}
    \centering
    \includegraphics[width=1\linewidth]{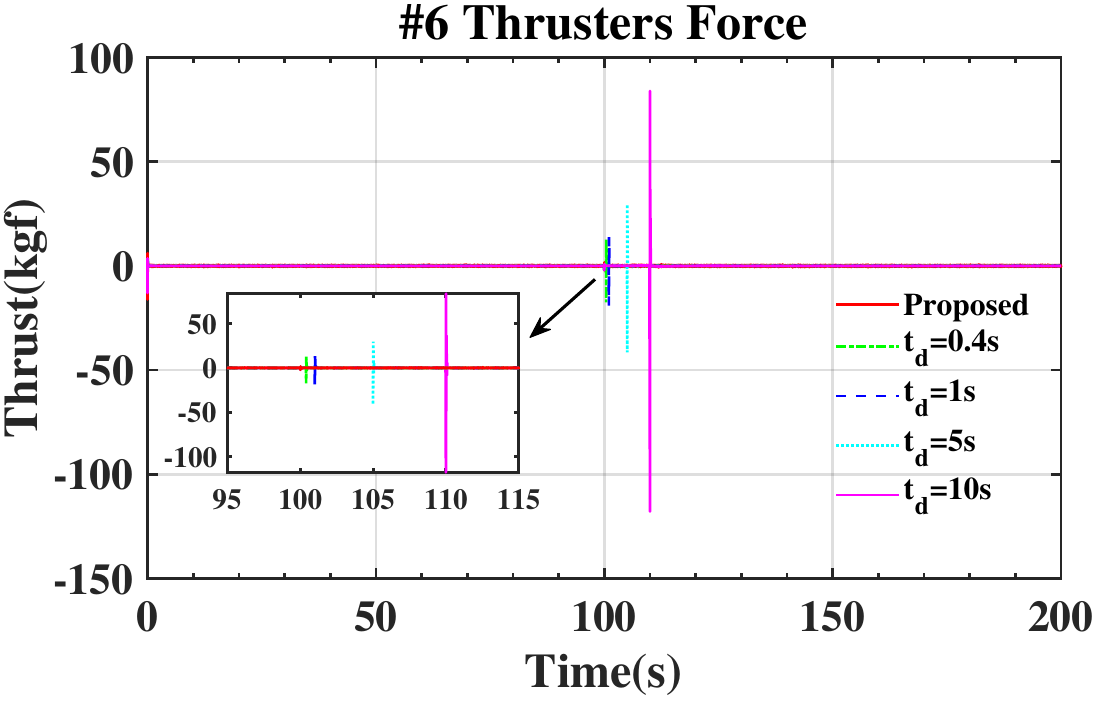}
    \label{fig:28}
\end{subfigure}\hfill
\begin{subfigure}{0.5\linewidth}
    \centering
    \includegraphics[width=1\linewidth]{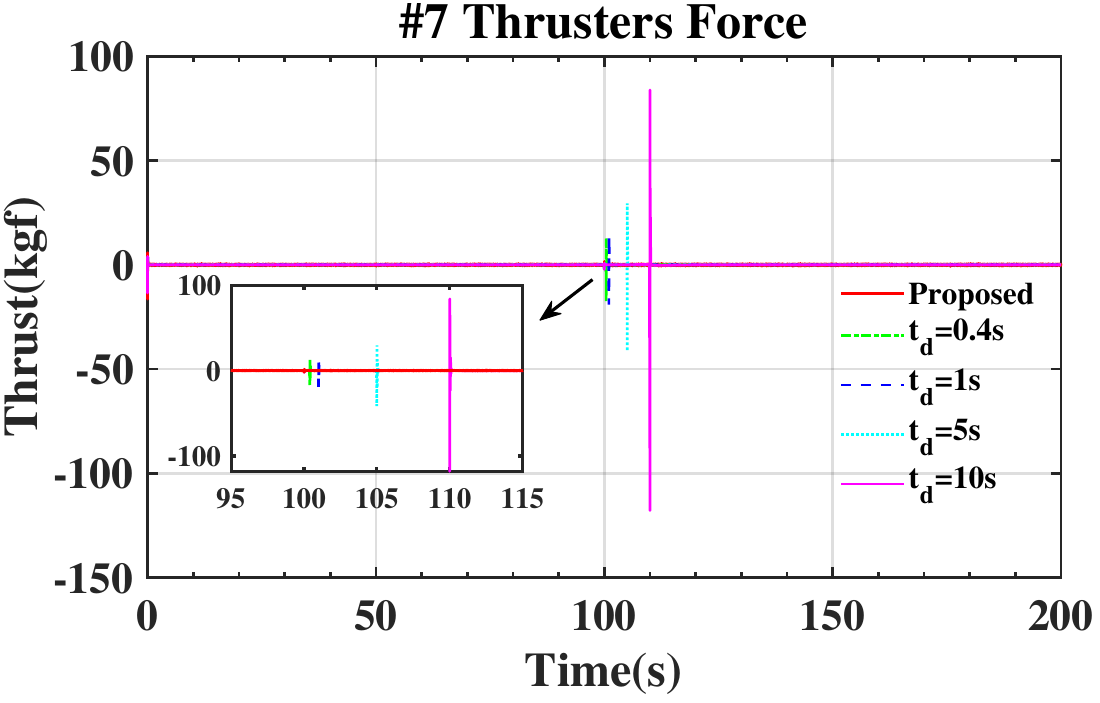}
    \label{fig:29}
\end{subfigure}\hfill
\begin{subfigure}{0.5\linewidth}
    \centering
    \includegraphics[width=1\linewidth]{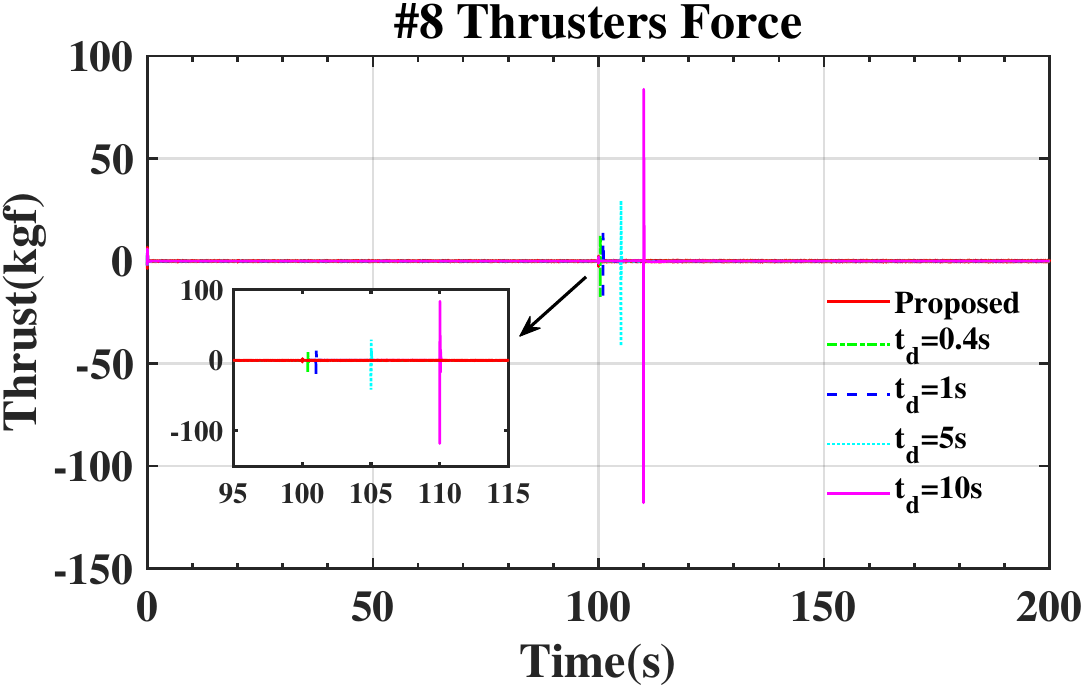}
\end{subfigure}
\caption{Comparison of thruster force from vertically arranged thrusters between the proposed fault tolerant control and hard-switch approach with different fault detection intervals of 0.1, 1, 5, and 10 seconds.}\label{fig:30}
\end{figure}

\Cref{fig:30} shows the response of the four vertically arranged thrusters when the thruster failure type shifts. Though Thrusters 2 and 4 theoretically do not influence the AUV motion and control in the depth direction, the failure switching significantly disturbs the AUV motion where it diverges from the reference trajectory. Therefore, adjustments in vertically arranged thrusters are necessary to maintain the trajectory tracking. The results in \Cref{fig:30} clearly demonstrates that the proposed adaptive fault-tolerant control method achieves the minimum adjustment in AUV thruster force, indicating minimized control cost in the control in case of failure type shift.
\clearpage

\section{Conclusions}\label{sec-conclusions}

This paper presents an adaptive fault-tolerant control for AUV trajectory tracking subject to multiple thruster failures and failure type shift. We construct a parameter space that encapsulates potential failure scenarios, and we adopt Bayesian approach in identifying the AUV failures. We derive the control law where we dynamically synthesizes control inputs from candidate failure models. Unlike conventional model-dependent fault-tolerant control, our proposed approach requires no prior knowledge about AUV internal dynamics, instead, it derives optimal control laws directly from system measurements. Furthermore, we address the lock-in issue in thruster failure type identification. We design a residual-triggered resetting mechanism which enables continuous and adaptive failure model update without compromising AUV trajectory tracking. We envision future work to validate the robustness of the proposed fault-tolerant control, and the examination of its scalability in complex marine missions with time-varying thruster constraints.



\section*{References}
\bibliographystyle{IEEEtran}
\bibliography{reference}

\end{document}